\documentclass[wcp]{jmlr}

\usepackage{longtable}
\usepackage{lscape}  
\usepackage{booktabs}

\usepackage{graphicx}
\usepackage{subcaption}
\usepackage{caption}
\usepackage{float} 
\usepackage{mathtools}%
\usepackage{setspace}

\makeatletter
\let\@jmlrpages\@empty 
\makeatother

\title[Best Arm Set Identification with Dual Constraints]{Risk-Averse Best Arm Set Identification with\\ Fixed
Budget and Fixed Confidence}

 \author{\Name{Shunta Nonaga} \Email{nonaga0811@eis.hokudai.ac.jp}\\
  \Name{Koji Tabata}$^*$ \Email{ktabata@es.hokudai.ac.jp}\\
  \Name{Yuta Mizuno} \Email{mizuno@es.hokudai.ac.jp}\\
  \Name{Tamiki Komatsuzaki}$^*$ \Email{tamiki@es.hokudai.ac.jp}\\
  \addr Hokkaido University, Hokkaido, Japan
  }

\begin{document}

\maketitle

\begin{abstract}
Decision making under uncertain environments in the maximization of expected reward while minimizing its risk is one of the ubiquitous problems in many subjects. Here, we introduce a novel problem setting in stochastic bandit optimization that jointly addresses two critical aspects of decision-making: maximizing expected reward and minimizing associated uncertainty, quantified via the \textit{mean-variance}(MV) criterion. Unlike traditional bandit formulations that focus solely on expected returns, our objective is to efficiently and accurately identify the Pareto-optimal set of arms that strikes the best trade-off between expected performance and risk. We propose a unified meta-algorithmic framework capable of operating under both fixed-confidence and fixed-budget regimes, achieved through adaptive design of confidence intervals tailored to each scenario using the same sample exploration strategy. We provide theoretical guarantees on the correctness of the returned solutions in both settings. To complement this theoretical analysis, we conduct extensive empirical evaluations across synthetic benchmarks, demonstrating that our approach outperforms existing methods in terms of both accuracy and sample efficiency, highlighting its broad applicability to risk-aware decision-making tasks in uncertain environments. 
\end{abstract}
\begin{keywords}
Stochastic multi-armed bandits; Multi-objective optimization; Pareto set identification
\end{keywords}

\section{Introduction}\label{sec1}
Stochastic multi-armed bandit (MAB) problems \cite{lattimore2020bandit} have emerged as a fundamental framework for online decision making under uncertainty, with broad applications ranging from adaptive drug discovery to recommendation systems \cite{madhukar2017new,qin2014contextual,li2010contextual,li2011unbiased}. The focus in MAB has been on the maximization of cumulative rewards by sequentially choosing from a set of options ---referred to as ``arm"--- based on stochastic feedback.  A conceptually distinct but equally fundamental variant within this framework is the best arm identification (BAI) problem, where the objective is not reward maximization over time, but rather the accurate identification of the optimal arm(s) using as few samples as possible. Because only the final decision matters, BAI operates under a pure exploration regime. This leads to unique algorithmic and theoretical challenges not encountered in classical reward maximization.
 
 BAI problems have been studied primarily under two canonical settings: (i) the fixed-confidence setting, where the goal is to guarantee the correctness of the identified arm(s) with high probability (at least $1 - \delta$ for any $\delta \in (0,1)$); and (ii) the fixed-budget setting, in which a learner is restricted to a fixed number of samples $T$ and must maximize the probability of correct identification.
Foundational work in the fixed-confidence setting introduced PAC-style guarantees
\cite{even2002pac}, later refined through approaches such as 
LUCB~\cite{kalyanakrishnan2012pac} achieving tighter bounds on sample complexity by leveraging confidence intervals. In the fixed-budget setting, algorithms such as Successive Rejects (SR)~\cite{audibert2010best} and Sequential Halving (SH)~\cite{karnin2013almost} eliminate suboptimal arms based on empirical ranking procedures. Despite their algorithmic differences, both settings share underlying complexity measures, such as the problem-dependent hardness parameter, 
often defined as the sum of inverse squared gaps between the best and suboptimal arms.

Recognizing the algorithmic parallels between the fixed-confidence and fixed-budget settings,~\cite{gabillon2012best} introduced the Unified Gap-based Exploration (UGapE) algorithm, which provides a single arm selection strategy applicable to both settings. This work laid the foundation for the unified algorithm design across different settings and emphasized the role of gap-based strategies in pure exploration.

More recently, increasing attention has been paid to \textit{risk-aware} variants of BAI, motivated by applications in medical trials or finance, where expected reward alone is insufficient \cite{huo2017risk,tamkin2019distributionally,keramati2020being,du2021continuous,chen2022strategies,shen2022risk}. In such variants, measure of variability, such as 
\emph{variance}, \emph{tail risk}, or \emph{quantiles}, must be taken into account in the decision process. For example,~\cite{hou2023almost} proposed the Variance-Aware (VA)-LUCB algorithm, which aims to identify the arm with the highest mean subject to a strict upper bound on variance. Their approach introduces a variance-aware hardness measure 
and shows nearly optimal sample complexity. Other approaches have explored alternative risk criteria such as Conditional Value-at-Risk (CVaR) and quantiles \cite{david2016pure}.

Parallel to this, much attention has been gained to incorporation with multi-objective BAI problems, where arms are evaluated across multiple criteria. Under the \textit{Pareto Set Identification} (PSI) framework, the goal is to identify the set of non-dominated arms (=Pareto-optimal arms) that are not outperformed across all objectives by any other. Early PSI algorithms such as the confidence-bound-based method proposed by~\cite{auer2016pareto} were developed under fixed-confidence settings, using uniform sampling and acceptance-rejection schemes. More recently, the adaptive LUCB-like algorithm for PSI~\cite{kone2023adaptive} has improved the sample efficiency by exploiting gap information.

Despite this progress, PSI under fixed-budget constraints has remained comparatively underexplored until recently.~\cite{kone2024fixedbudget} introduced the \textit{Empirical Gap Elimination (EGE)} framework, which generalizes SR and SH to the multi-objective setting. EGE estimates empirical gaps to eliminate arms and classifies them as Pareto-optimal or suboptimal. The EGE-based algorithms, EGE-SR and EGE-SH, were found to achieve exponential decay in error probability with respect to budget and are near-optimal according to an information-theoretic lower bound.  
Despite recent progresses, some limitations yet remain in the multi-objective BAI literature. Foremost among these is the limitation on both theoretical unification and practical applicability across both fixed-confidence, and fixed-budget canonical settings: most existing algorithms are tailored specifically to either of the two settings. In parallel, 
although recent efforts have been devoted in introducing risk-awareness into BAI ---such as VA-LUCB under variance constraints--- these approaches typically handle risk in isolation, without integrating it into multi-objective frameworks like Pareto Set Identification (PSI)~\cite{ulrich2008pattern,kone2025bandit}. Moreover, existing PSI algorithms either ignore uncertainty (risk) altogether or treat it as an independent objective, lacking a principled scheme to jointly evaluate utility and risk in arm selection.

To address these challenges, we propose a novel multi-objective optimization framework to take into account both mean and risk simultaneously that bridges the fixed-confidence and fixed-budget paradigms through a unified arm selection strategy, modulated only by confidence intervals and setting-specific stopping rules, we call RAMGapE (Risk-Averse Multi-objective Gap-based Exploration). Central to RAMGapE is a new gap-based criterion that incorporates both the expected reward and the associated risk, quantified through a mean-variance trade-off. This allows for efficient identification of $\epsilon$-Pareto optimal arms while explicitly accounting for risk. Our theoretical analysis provides guarantees on correctness and sample complexity, while extensive experiments demonstrate that RAMGapE significantly outperforms existing methods in risk-sensitive settings—achieving superior decision quality with fewer samples. Specifically, the novelty of our approach lies in overcoming the limitations of prior studies through a new theoretical framework, summarized in three key aspects:
\begin{enumerate}
    \item [1]
    Adaptation of Gap-Based Analysis to Partial Orders: Unlike traditional methods that often assume a total order of arms (e.g., ranked by their mean rewards), our algorithm extends gap-based analysis to a partial-order setting defined by mean-risk Pareto dominance. 
    \item [2]
    Handling of Variable-Size Pareto Sets: We remove the common assumption of a fixed number of optimal arms or a pre-defined boundary arm for comparison. RAMGapE is designed to identify the entire Pareto set whose size is unknown a priori and can vary, making it applicable to a wider range of real-world problems.
    \item [3] 
    A Novel Exploration Rule Targeting Pareto Dominance: We introduce a new exploration strategy that explicitly targets the structure of Pareto dominance. Instead of focusing on a single best arm, the algorithm efficiently allocates samples to resolve uncertainties along the Pareto frontier, pruning provably suboptimal arms and identifying the set of non-dominated solutions.
\end{enumerate}
To our knowledge, RAMGapE is the first algorithm to present provable guarantees, having the above features, for multi-objective, risk-averse PSI whose two objective variables are dependent to each other via the same reward distribution in both fixed-confidence and fixed-budget frameworks.

\section{Problem Setting}\label{sec2}

In this section, we introduce the definitions and notation used throughout this paper. Let $[K]=\{1,2,\ldots,K\}$ be the set of arms such that each arm $i\in[K]$ is characterized by a reward distribution $\nu_i$ bounded in $[0,1]$ with mean $\mu_i$ and variance $\sigma^2_i$. Here we employ a risk criteria based on \textit{mean-variance} (MV)~\cite{sani2012risk} as the second co-equal objective. 
The smaller the value of MV, the lower the risk of the arm. 
Here, $\rho~(\geq0)$ is a hyperparameter to control the weight of its risk in the search, that is, when $\rho\rightarrow\infty$, the minimization of MV corresponds to finding arm(s) with a larger mean(s) without taking care of its variance, while it corresponds to finding arm(s) with smaller variance(s) when $\rho \rightarrow 0$  ($\rho$ has the dimension of mean $\mu_i$). 
We introduce a parameter $\alpha~(>0)$ to scale the MV measure as 
$\xi_i\coloneq\alpha\text{MV}_i$ for each arm $i$, which preserves the relative position of the risk measure.
%
Later, we set $\alpha = \frac{1}{3 + \rho}$ to allow the construction of confidence intervals with equal widths for both mean and risk measures. 
In classical formulation, variance captures reward uncertainty. However, when identifying
Pareto-optimal solutions over mean and variance as risk measure, arms with very low mean but
low variance can still be deemed Pareto-optimal, even though they are of little practical
importance. Using mean-variance (MV) criterion with adjusting the parameter
$\rho$, we can design the identification of Pareto solutions that have both relatively high mean and low risk. 
However, due to its second-order moment property, MV accounts for the ``risk'' symmetrically. When the underlying reward distribution is skewed and higher-order moments are non-zero, MV may not adequately capture the severity of rare events (tail risk). Among other risk measures, for example, Conditional Value-at-Risk (CVaR)~\cite{rockafellar2000optimization} is another possible choice of risk measure. CVaR quantifies the expected loss in the worst-case scenarios. 
The challenge of extending our gap-based framework to asymmetric, tail-focused risk measures like CVaR remains one of the forthcoming subjects to be resolved, as discussed in Section~\ref{sec:conclusion}.

Next, we address Pareto optimality when the two stochastic variables $\mu_i$ and $\xi_i$ are used as objective criteria.

We say that arm \(j\) \textit{strictly dominates} arm \(i\), denoted as \(j \succ i\), if both \(\mu_j > \mu_i\) and \(\xi_j < \xi_i\) hold; that is, arm \(j\) has a strictly higher expected reward and strictly lower risk than those of arm \(i\). An arm \(i \in [K]\) is said to be \textit{Pareto optimal} if there exists no other arm \(j \in [K]\) such that \(j \succ i\). In other words, arm \(i\) is not strictly dominated by any other arm(s). We denote by \(D^+\) the set of all arms that satisfy this Pareto optimality condition.

\noindent
For each arm $i\in[K]$, according to ~\cite{kone2024fixedbudget}, we define a gap $\Delta_i$ as
\begin{equation}\Delta_i\coloneqq
\left\{
\begin{matrix}
  \min\left\{\underset{j\in D^+\setminus \{i\}}{\operatorname{min}}\
  \Big(\min({M}(i,j),{M}(j,i))\Big),\underset{j\notin D^+}{\min}\ ({M}(j,i)^++\Delta_j)\right\}& \text{if }i\in D^+; \\
  \underset{j\in D^+\text{ s.t. }j\succ i}{\operatorname{max}}\ {m}(i,j) & \text{if } i\notin D^+,
\end{matrix}
\right.
\label{eq:def_gap}
\end{equation}
where $m(i,j)\coloneq\min(\mu_j-\mu_i,\xi_i-\xi_j)$, $M(i,j)\coloneq\max(\mu_i-\mu_j,\xi_j-\xi_i)$, ${M}(j,i)^+\coloneq\max({M}(j,i),0).$ 

The definition of gap tells us that for $i\in D^+$, $\Delta_i$ properly quantifies how well arm $i$ separates itself from other arms, capturing both the minimal margin from non-Pareto arms and the proximity to other Pareto-optimal arms,
and for $i\notin D^+$, $\Delta_i$ does to what degree the non-Pareto optimal arm $i$ is dominated by the other arms at most. We illustrate these quantities and explain the details in
Appendix \ref{illust_gap}. Given an allowance $\epsilon>0$ defined by a user, a subset $S\subseteq[K]$ is called a $\epsilon$-Pareto set if it satisfies the following condition,
\begin{align*}
    \forall{i}\in S, \, \forall{j}\in[K], \, \mu_i > \mu_j-\epsilon \lor \xi_i < \xi_j+\epsilon, \\
    \forall{i}\not\in S, \, \exists{j}\in[K], \, \mu_i \leq \mu_j - \epsilon \land \xi_i \geq \xi_j + \epsilon.
\end{align*}
Hereinafter, we formulate the 
Risk-Averse Best Arm Set Identification Problem 
as the problem to find an $\epsilon$-Pareto set for expected means and their risks. Note that $\epsilon$ can be regarded as a resolution associated with the observation in question and $\epsilon\rightarrow 0$ converges to the problem without any error in measurement. 
\vskip\baselineskip
\noindent
The Risk-Averse Best Arm Set Identification Problem can be formalized as a process between a stochastic bandit environment and a forecaster. The reward distributions $\{\nu_i\}_{i=1}^K$ inherent to each arm are unknown a priori to the forecaster. At each round $t$, the forecaster pulls an arm $I(t)\in[K]$ and observes a sample independently drawn from the identical distribution $\nu_{I(t)}$. Let $T_i(t)$ be the number of times that arm $i$ has been pulled up to the round $t$, the forecaster estimates the expected value of mean, variance, and risk of this arm by $\hat{\mu}_i(t)=\frac{1}{T_i(t)}\sum_{s=1}^{T_i(t)}X_i(s)$, $\hat{\sigma}^2_i(t)=\hat{\mu}_i^{(2)}(t)-\hat{\mu}_i^2(t)$, and $\hat{\xi}_i(t)=\alpha(\hat{\sigma}^2_i(t)-\rho\hat{\mu}_i(t))$, where $X_i(s)$ and $\hat{\mu}_i^{(2)}(t)$ are the $s$-th sample observed from $\nu_i$ and $\frac{1}{T_i(t)}\sum_{s=1}^{T_i(t)}X_i^2(s)$, respectively. For any set $S\subseteq[K]$ and arm $i\in[K]$, 
we introduce the notations of arm simple regret $r_i(S)$ for arm $i$ as well as (set) simple regret $r_S$  for set of arms $S$ as follows:  
\begin{eqnarray}
    r_i(S) &=& \begin{cases} 
  \Delta_i& \text{if }i\in S\bigtriangleup{D^+} \\
  0 & \text{otherwise} \end{cases},\\
r_S&=& \max_{i \in [K]} r_i(S) 
\end{eqnarray}
where $A \bigtriangleup B := (A \setminus B) \cup (B \setminus A)$ for any sets $A$ and $B$. 
\vskip\baselineskip
\noindent
We define a temporary set of Pareto arms with respect to the empirical values at round $t$ as
\[
\widehat{D}^+_t \coloneq \left\{ i \in [K] \;\middle|\; \forall j \in [K],\ j \not\succ_t i \right\},
\]
where the empirical (strict) dominance relation $\succ_t$ is defined as follows:

\begin{definition}[Empirical Dominance Relation]
For any two arms $i, j \in [K]$ in round $t$, we say that arm $j$ \textit{strictly dominates} arm $i$ at round $t$, denoted by $j \succ_t i$, if
\[
\hat{\mu}_j(t) > \hat{\mu}_i(t) \quad \text{and} \quad \hat{\xi}_j(t) < \hat{\xi}_i(t).
\]
In other words, arm $j$ is better than arm $i$ in both mean and risk estimates at round $t$.  
We denote $j \not\succ_t i$ if this condition does not hold.
\end{definition}

The simple regret in each round $t$ can be written as $r_{\widehat{D}^+_t}$.  Returning an $\epsilon$-Pareto set is then equivalent to having $r_{\widehat{D}^+_t}$ smaller than $\epsilon$. Given an allowance $\epsilon$, we formalize the two settings of fixed budget and fixed confidence.
\vskip\baselineskip
\noindent
\textbf{Fixed budget.} The objective is to return the set of $\epsilon$-Pareto arms with the highest possible confidence level using a fixed budget of $n$ rounds. Formally, given a budget $n$, the performance of the forecaster is measured by the probability $\widetilde{\delta}$ of not satisfying the conditions of the set of $\epsilon$-Pareto arms, i.e., $\widetilde{\delta}=\mathbb{P}\Big[r_{\widehat{D}^+_n}\geq \epsilon\Big]$, the smaller $\widetilde{\delta}$, the better the algorithm.
\vskip\baselineskip
\noindent
\textbf{Fixed confidence.} The objective is to design a forecaster that stops as soon as possible and returns the set of $\epsilon$-Pareto arms with fixed confidence. Let $\widetilde{n}$ be the round at which the algorithm stops, and let $\widehat{D}^+_{\widetilde{n}}$ be the set of arms returned. Given a fixed confidence $\delta$, the forecaster must guarantee that $\mathbb{P}\Big[r_{\widehat{D}^+_{\widetilde{n}}}\geq\epsilon\Big]\leq\delta$. The forecaster performance is evaluated at the stopping round $\widetilde{n}$.
\vskip\baselineskip
\noindent
Although traditionally treated as distinct problems, in Section \ref{sec3} we present a unified arm selection strategy that applies to both settings, differing only in the choice of stopping criterion.

\section{Risk-Averse Multi-objective 
 Gap-based Exploration Algorithm}\label{sec3}

In this section, we present the risk-averse gap-based exploration algorithm (RAMGapE) meta-algorithm, involving its implementation for 
fixed budget and fixed confidence settings, named RAMGapEb and RAMGapEc, respectively. The algorithm in each 
setting uses a common arm selection strategy, PullArm (Algorithm \ref{alg:pullarm}) (see also the pseudo-code Algorithm \ref{alg:ramgape}). 
RAMGapEb and RAMGapEc
return an $\epsilon$-Pareto set using the same definition of temporal Pareto set 
$\widehat{D}$.
They 
only differ in the stopping rule. Given an allowance $\epsilon$, both algorithms 
first suppose constant parameters such as the budget $n$ and the hyperparameter $a$ that controls the exploration rate RAMGapEb, the confidence level $\delta$ in RAMGapEc, respectively. RAMGapEb runs for $n$ rounds and returns a set of arms 
$\widehat{D}^+_n$
, whereas RAMGapEc runs until it achieves the required confidence level $\delta$ so that the probability of correctly extracting the Pareto optimal set is greater than $1-\delta$ under the given allowance 
$\epsilon$.
The difference is caused by the different objectives of the two algorithms: RAMGapEb aims to maximize the quality of prediction under the fixed budget but RAMGapEc aims to minimize budget required to accomplish the given fixed confidence level.
\begin{algorithm}[tb]
\rule{\linewidth}{1pt}
\caption{PullArm}\label{alg:pullarm}
\vspace{-1.6ex}
\rule{\linewidth}{0.5pt}

\KwIn{$t$, $\{T_i(t)\}_{i=1}^K$, $\{\beta_i(t)\}_{i=1}^K$, $\{\hat{\mu}_i(t)\}_{i=1}^K$, $\{\hat{\xi}_i(t)\}_{i=1}^K$}

\If{$\exists i$ such that $T_i(t) \le 2$}{
    \Return{$\arg\min_{i \in [K]} T_i(t)$}
}
Compute $V_i(t)$ for each arm $i \in [K]$\;

Determine $m_t$ and $p_t$ using Eq.~\ref{mt} and Eq.~\ref{pt}\;

\Return{$\arg\max_{i \in \{m_t, p_t\}} \beta_i(t)$}

\vspace{-1.5ex}
\rule{\linewidth}{1pt}
\end{algorithm}
\begin{algorithm}[tb]
\rule{\linewidth}{1pt}
\caption{RAMGapE}\label{alg:ramgape}
\vspace{-1.6ex}
\rule{\linewidth}{0.5pt}

\KwIn{$K$, $a$, $n$, $\epsilon$, $\rho$}

Initialize $T_{i}(1) \gets 0$, $\beta_{i}(1) \gets 0$, $\hat{\mu}_{i}(1) \gets 0$, $\hat{\xi}_{i}(1) \gets 0$ for $i = 1, 2, \ldots, K$\;

Set $t \gets 1$\;

\While{$t \le n$}{
    $I(t) \gets \text{PullArm}\left(t, \{T_{i}(t)\}, \{\beta_{i}(t)\}, \{\hat{\mu}_{i}(t)\}, \{\hat{\xi}_{i}(t)\} \right)$\;
    
    Observe $X_{I(t)}(T_{I(t)}(t)+1) \sim \nu_{I(t)}$\;
    
    $t \gets t + 1$\;

    Update $\hat{\mu}_{I(t)}(t)$, $\hat{\xi}_{I(t)}(t)$, $\beta_{I(t)}(t)$, and $T_{I(t)}(t)$\;

    \tcp*[l]{(RAMGapEb)} 
    \If{$t > n$}{\textbf{break}}

    \tcp*[l]{(RAMGapEc)} 
    \If{$t > 2K\wedge V(t) < \epsilon$}{\textbf{break}}
}
\Return{$\widehat{D}^+_n$}

\vspace{-1.5ex}
\rule{\linewidth}{1pt}
\end{algorithm}

\vskip\baselineskip
\noindent
To initialize variance estimation, each arm is first pulled twice before the adaptive exploration begins. This initialization step guarantees that variance estimates are properly defined when computing the risk-based criteria used throughout the algorithm. In PullArm (Algorithm \ref{alg:pullarm}),
at each round $t$ and for each arm $i\in[K]$, RAMGapE first uses the information observed up to the round $t$ and computes quantities $V_i(t),V(t),m_t,p_t,$ and $I(t)$ that are defined by 
\begin{eqnarray}\label{Vi(t)}V_i(t)&\coloneqq&
\left\{
\begin{matrix}
  \underset{j\neq i}{\operatorname{max}}\ \min\Big(\overline{\mu}_j(t)-\underline{\mu}_i(t),\overline{\xi}_i(t)-\underline{\xi}_j(t)\Big)& \text{if }i\in \widehat{D}^+_t \\
  \underset{j\in\widehat{D}^+_t \text{ s.t. }j\underset{t}{\succ} i}{\operatorname{min}}\ \max\Big(\overline{\mu}_i(t)-\underline{\mu}_j(t),\overline{\xi}_j(t)-\underline{\xi}_i(t)\Big) & \text{if } i\notin \widehat{D}^+_t
\end{matrix}
\right.,\\
V(t)&\coloneq&\underset{i\in[K]}{\max}\ V_i(t),\\
\label{mt}m_t&\coloneq&\underset{i\in[K]}{\operatorname{argmax}}\ V_i(t)\\
\label{pt}p_t&\coloneqq&
\left\{
\begin{matrix}
  \underset{j\neq m_t}{\operatorname{argmax}}\ \min\Big(\overline{\mu}_j(t)-\underline{\mu}_{m_t}(t),\overline{\xi}_{m_t}(t)-\underline{\xi}_j(t)\Big)& \text{if }m_t\in \widehat{D}^+_t \\
  \underset{j\in\widehat{D}^+_t \text{ s.t. }j\underset{t}{\succ}m_t}{\operatorname{argmin}}\ \max\Big(\overline{\mu}_{m_t}(t)-\underline{\mu}_j(t),\overline{\xi}_j(t)-\underline{\xi}_{m_t}(t)\Big) & \text{if } m_t\notin \widehat{D}^+_t
\end{matrix}
\right..
\end{eqnarray}
Here $\overline{\mu}_i(t), \underline{\mu}_i(t),\overline{\xi}_i(t)\text{ and } \underline{\xi}_i(t)$ represents 
the upper and lower bounds of the mean ($\mu_i$) and risk ($\xi_i$) of the arm $i$ after $t$ rounds, respectively. In brief, $V_i(t)$ estimates the maximum gap of arm $i$ from the rest by comparing the pessimistic predictions of $\mu_i$ and $\xi_i$ and the optimistic predictions of those of the other arms when $i$ belongs to the temporal Pareto set $\widehat{D}^+_t$ defined by the sample mean and sample risk at round $t$. Likewise, when $i$ does not belong to $\widehat{D}^+_t$, it estimates the minimum gap between $(\mu_i,\xi_i)$ of the arm $i$ and those of the arms belonging to $\widehat{D}^+_t$ to dominate the arm $i$. These quantities are defined by
\begin{equation}\label{bound}
    \forall i\in[K], \forall t,\ \left\{
\begin{matrix}
  \overline{\mu}_i(t)\coloneq\hat{\mu}_i(t)+\beta_i(t)\\
  \underline{\mu}_i(t)\coloneq\hat{\mu}_i(t)-\beta_i(t)
\end{matrix}
\right.\ ,\
\left\{
\begin{matrix}
  \overline{\xi}_i(t)\coloneq\hat{\xi}_i(t)+\beta_i(t)\\
  \underline{\xi}_i(t)\coloneq\hat{\xi}_i(t)-\beta_i(t)
\end{matrix}
\right.,
\end{equation}
where 
$\beta_i(t)$ denotes their
confidence intervals and a parameter denoted by $a$ was employed in the definition of $\beta_i$, whose shape strictly depends on the concentration bound used by the algorithm. For example, we can derive $\beta_i$ from the Hoeffding-Azuma inequality ~\cite{azuma1967weighted,tropp2012user} as
\begin{equation}\label{confidence_bound}
\begin{matrix}
 \text{ RAMGapEb: }\beta_i(t)=&\sqrt{\frac{a}{T_i(t)}}, \\ \text{ RAMGapEc: }\beta_i(t)=&\sqrt{\frac{4}{T_i(t)}\ln\frac{8K(\log_2 T_i(t))^2}{\delta}}.
\end{matrix}
\end{equation}
We introduce a quantity $V_S(t)$ for a set $S$ as $V_S(t)  \coloneq\max_{i\in S} V_i(t)$. After computing the quantities for all arms, RAMGapE selects two key arms:  $m_t$, the arm with the largest $V_i(t)$, and $p_t$, the most relevant comparison arm to $m_t$ based on the dominance relation. Depending on whether $m_t$ and $p_t$ are included in the currently estimated temporal Pareto set $\widehat{D}^+_t$, these arms may represent potentially optimal or suboptimal candidates. If both are in $\widehat{D}^+_t$, they are regarded as highly uncertain arms that could be Pareto optimal. RAMGapE then selects the arm with a fewer number of pulls between $m_t$ and $p_t$, thereby prioritizing exploration toward the arm with greater uncertainty. The algorithm pulls the selected arm, observes its reward, and updates its empirical mean $\hat{\mu}_i(t)$, risk estimate $\hat{\xi}_i(t)$, and pull count $T_i(t)$.

The core mechanism of RAMGapE is distribution-agnostic, requiring only valid confidence intervals, $\beta_i(t)$. While we employ Hoeffding-based bounds for Beta distributions, this choice can be adapted. For instance, bounds for sub-Gaussian variables are applicable for unbounded rewards; we validate this empirically in Appendix~\ref{sec:gaussian}. For heavy-tailed rewards, robust estimators like truncated means~\cite{bubeck2013bandits} or Catoni's M-estimator~\cite{catoni2012challenging} can be readily integrated to ensure valid concentration bounds without altering the algorithm's main structure. This modularity confirms the broad applicability of RAMGapE.

\subsection*{Theoretical Guarantees}
We provide theoretical guarantees for RAMGapE under both the fixed-confidence and fixed-budget settings. The core of our analysis relies on establishing a high-probability event $\mathcal{E}$ (see Eq.~\ref{event} in Appendix~\ref{theoretical_analysis}) where all empirical estimates remain within their confidence intervals. Under this event, we can guarantee the algorithm's performance. The detailed proofs for the following theorems are given in Appendix \ref{theoretical_analysis}.
\begin{theorem}
(Error Bound for Fixed-Budget) If we run RAMGapEb with parameter $0<a\leq\frac{n-2K}{16K}\epsilon^2$ for rounds $n$,  total number of arms $K$ and allowance rate $\epsilon$, its simple regret $r_{\widehat{D}^+_n}$ satisfies
\begin{equation}
    \widetilde{\delta}=\mathbb{P}\left[r_{\widehat{D}_n}\geq\epsilon\right]\leq4Kn\exp(-2a),\nonumber
\end{equation}    
and, in particular, this probability is minimized at $a=\frac{n-2K}{16K}\epsilon^2$.
\end{theorem}
\begin{theorem}
    (Correctness and Termination for Fixed-Confidence) The RAMGapEc algorithm stops after $\widetilde{n}$ rounds and returns an $\epsilon$-Pareto set, $\widehat{D}^+_{\widetilde{n}}$, that satisfies
\begin{equation*}
    \mathbb{P}\left[r_{\widehat{D}^+_{\widetilde{n}}}\leq\epsilon\wedge\widetilde{n}\leq N\right]\geq1-\delta,
\end{equation*}
where $N=2K+ \mathcal{O}\left( \frac{K}{\epsilon^2} \log\left( \frac{K \log_2^2(1/\epsilon)}{\delta} \right) \right)$ with confidence level $\delta~(<1)$.
\end{theorem}

\section{Experiments}

In this section, we evaluate the performance of the proposed algorithm RAMGapE under both the fixed-confidence and fixed-budget settings. We compare the performance with those of other algorithms, including the standard Round-Robin strategy and several previously proposed approaches for risk-averse and multi-objective bandit problems.

\subsection*{Fixed-Confidence Setting}

In the fixed-confidence setting, we compare RAMGapE with the following three representative algorithms. Round-Robin uniformly samples each arm and serves as a fundamental baseline. Dominated Elimination Round-Robin (DE Round-Robin) (also used as a baseline in~\cite{kone2024fixedbudget}) improves upon this by eliminating empirically dominated arms based on observed values. Risk-Averse LUCB (RA-LUCB) extends the classical LUCB algorithm \cite{kalyanakrishnan2012pac} to risk-sensitive settings, where it selects and pulls two arms—denoted $m_t$ and $p_t$—in each round (see also pseudo-codes in Appendix \ref{comparison_methods}). In contrast, RAMGapE differs from RA-LUCB in that it pulls only the less frequently sampled of the two arms $m_t$ and $p_t$. This leads to improved sample efficiency while maintaining identification accuracy, and this selection rule forms the main distinction between the two algorithms.
\vskip\baselineskip\noindent
\textbf{Experiment 1 (Comparison of Stopping Time):}\\
\noindent
We compare the number of rounds required by each method to meet the stopping condition in 50 problem instances. The reward for each arm follows a Beta distribution, with means in $[0.4,0.6]$ and variances in $[0.01,0.2]$, and the number of arms is set to $K=10$ (see Table~\ref{table_50_pattarns}). The algorithmic parameters are fixed at $(\delta, \epsilon, \rho) = (0.05, 0.1, 0.01)$. Please see also the similar experiment with $\epsilon=0.05$ in Fig.\ref{fig:complexity_analysis_fig} (Appendix~\ref{sec:epsilon_change}).
\vskip\baselineskip\noindent
\textbf{Experiment 2 (Comparison of Confidence Intervals at Stopping Time):}\
\noindent
Using the same set of problem instances as in Experiment 1, we compare the width of confidence intervals at the stopping point for each algorithm. The tolerance parameter is set to $\epsilon=0$, enabling us to assess how conservative or aggressive each method is in its stopping criterion.
We consider two settings:
\textbf{Experiment 2.1} corresponds to instances where the number of Pareto-optimal arms is small (about arms set, see Table~\ref{table_50_pattarns}, pattern 10), while
\textbf{Experiment 2.2} targets instances where the number of Pareto-optimal arms is large (about arms set, see Table~\ref{table_50_pattarns}, pattern 46).
This allows us to evaluate the behavior of the algorithms under different levels of Pareto set complexity.

\begin{figure}[tb]
  \centering

  \subfigure[vs RA-LUCB]{%
    \includegraphics[width=0.325\textwidth]{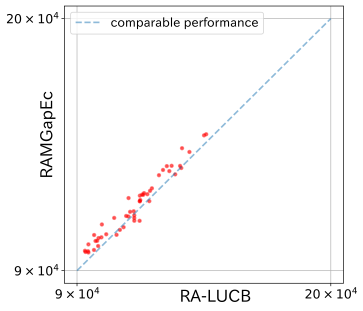}}
  \hfill
  \subfigure[vs DE Round-Robin]{%
    \includegraphics[width=0.325\textwidth]{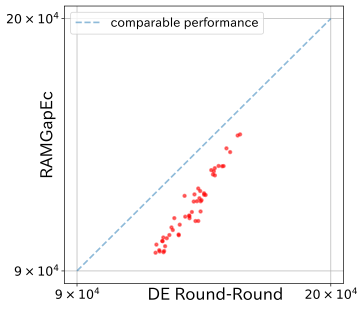}}
    \hfill
    \subfigure[vs Round-Robin]{%
    \includegraphics[width=0.325\textwidth]{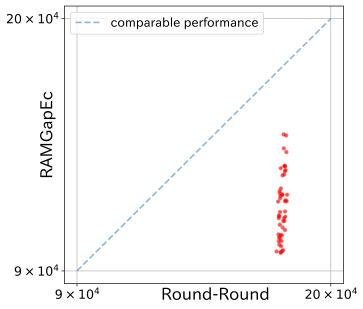}}
  \caption{\textbf{Stopping Time Comparison of Experiment 1 with $\epsilon=0.1$.} 
  The blue dashed line corresponds to the identity line, i.e., the set of points where both methods terminate at the same time, indicating comparable performance. Points located below this line signify that the proposed method stops earlier than the baseline.
  }
  \label{fig:experiment1}
\end{figure}

\begin{figure}[t]
  \centering

  \subfigure[RAMGapEc]{%
    \includegraphics[width=0.244\textwidth]{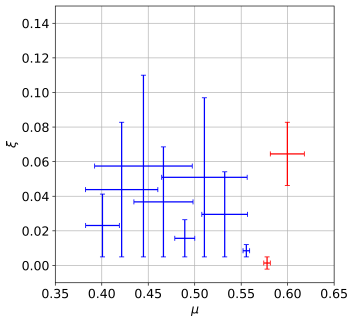}}
  \hfill
  \subfigure[RA-LUCB]{%
    \includegraphics[width=0.244\textwidth]{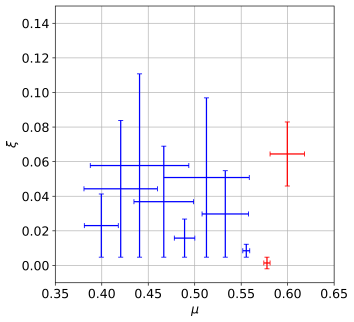}}
  \hfill
    \subfigure[DE Round-Robin]{%
    \includegraphics[width=0.244\textwidth]{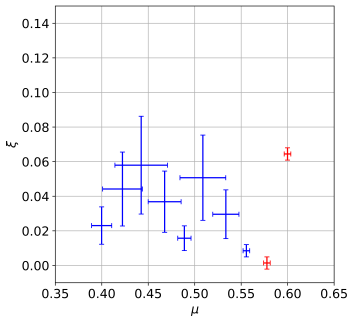}}
    \hfill
    \subfigure[Round-Robin]{%
    \includegraphics[width=0.244\textwidth]{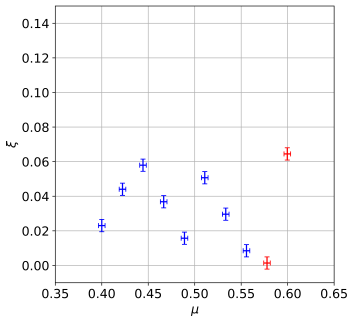}}
  \caption{\textbf{Visualization of confidence intervals at stopping time (Experiment 2.1).}
    Each panel shows the empirical mean (horizontal axis) and scaled risk (vertical axis: $\xi = \alpha(\sigma^2 - \rho \mu)$) of each arm at the termination round for different algorithms. 
    The crosses represent confidence intervals of each arm; the longer the arms of the crossed interval, the fewer the samples allocated to that arm.
    Red points (=crosses) indicate arms included in the returned set $\widehat{D}^+_t$, while blue points indicate excluded arms.
    RAMGapEc and RA-LUCB not only avoid over-sampling 
    non-Pareto arms (shown in blue), but also limit sampling for some arms included in $\widehat{D}^+_t$, particularly those located on the far right of the plot (i.e., arms with high mean but less impact on Pareto set boundaries). These arms exhibit wider confidence intervals, reflecting lower sample counts.
    This behavior highlights the algorithms’ ability to allocate samples efficiently, gathering just enough information for confident identification without unnecessary exploration.
    The total sample counts of these examples are:
    RAMGapEc: 9,697,292; RA-LUCB: 9,728,010; DE Round-Robin: 15,283,296; Round-Robin: 43,548,822.}
  \label{fig:experiment2_1}
\end{figure}



\begin{figure}[t]
  \centering

  \subfigure[RAMGapEc]{%
    \includegraphics[width=0.244\textwidth]{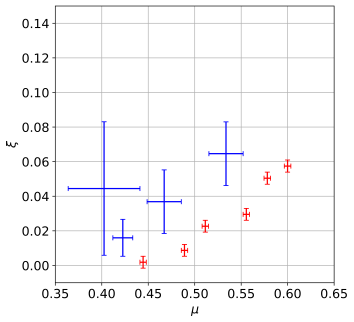}}
  \hfill
  \subfigure[RA-LUCB]{%
    \includegraphics[width=0.244\textwidth]{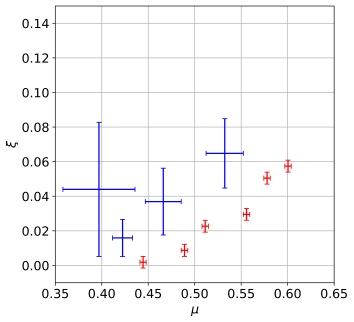}}
  \hfill
    \subfigure[DE Round-Robin]{%
    \includegraphics[width=0.244\textwidth]{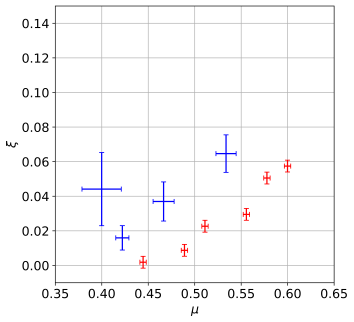}}
    \hfill
    \subfigure[Round-Robin]{%
    \includegraphics[width=0.244\textwidth]{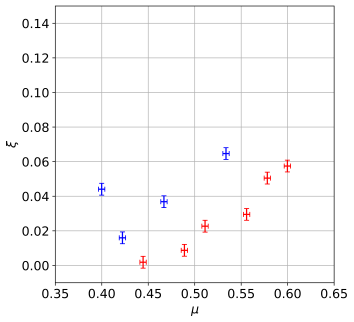}}
  \caption{\textbf{Visualization of confidence intervals at stopping time (Experiment 2.2).}
    The meanings of the crosses, and the colors are the same as in Fig.~\ref {fig:experiment2_1}. The total number of samples used for these examples are:
    RAMGapEc: 28,261,200; RA-LUCB: 28,486,332; DE Round-Robin: 30,041,447; Round-Robin: 46,905,293.}
  \label{fig:experiment2_2}
\end{figure}

\subsection*{Fixed-Budget Setting}

In the fixed-budget setting, we compare RAMGapE with several algorithms: the standard Round-Robin, Least-Important Elimination Round-Robin (LIE Round-Robin), RA-LUCB adapted for the fixed-budget case, the risk-sensitive $\xi$-LCB (used as a baseline in \cite{sani2012risk}), hypervolume-based HVI-Pareto method (see e.g.,~\cite{yang2019efficient,zitzler2007hypervolume,cao2015using} for the definition and applications of hypervolume), and the Empirical Gap-based Pareto Set Exploration (EGP) (see the relevance in \cite{kone2024fixedbudget}) (see the pseudo-codes of these comparison algorithms in Appendix \ref{comparison_methods}).

It should be noted that Round-Robin-based evaluation strategies have traditionally been used in domains such as medicine, where repeated sampling and fair treatment allocation are a common strategy (see, for example,~\cite{alzheimers,ovariancancer}). 
\vskip\baselineskip\noindent
\textbf{Experiment 3 (Comparison of Average Simple Regret with a Small Number of Arms):}\\
\noindent
We evaluate average simple regret for each method under $K=10$ arms (see Table~\ref{table_10_pattarns_b}). Each arm’s reward distribution is a Beta distribution with randomly sampled means and variances from 
$[0.4,0.6]$ and 
$[0.01,0.2]$, respectively. The parameter $a$ is set as $\frac{n-2K}{16K}\epsilon^2$ (see Appendix \ref{theorem:fb_regret_bound}). 
All algorithms are executed for 50 independent trials with $T=10,000$ rounds. In order to reduce the influence of outliers, the lower and upper 25\% of the simple regret values at each time round are excluded and the remaining middle 50\% of the simple regret values are averaged and used for the evaluation of the algorithm performance. 

\vskip\baselineskip\noindent
\textbf{Experiment 4 (Comparison of Average Simple Regret with a Large Number of Arms):}\\
\noindent
To assess scalability, we increase the number of arms to 
$K=100$ (see Table~\ref{table_100_pattarns_b}), while keeping the same 
settings as in Experiment 3.

In all experiments, we set the risk coefficient to $\alpha=\frac{1}{3+\rho}$, ensuring that the widths of confidence intervals for both the mean and the risk metric are balanced (see Appendix \ref{CI_derivation}). The use of Beta distributions allows us to model a variety of shapes—unimodal, U-shaped, monotonic, and uniform—making the evaluation more reflective of real-world scenarios. Further implementation details and algorithmic formulations are provided in Appendix \ref{comparison_methods}.



\begin{figure}[tb]
  \centering

  \subfigure[Experiment 3 with $K = 10$ arms]{%
    \includegraphics[width=0.45\textwidth]{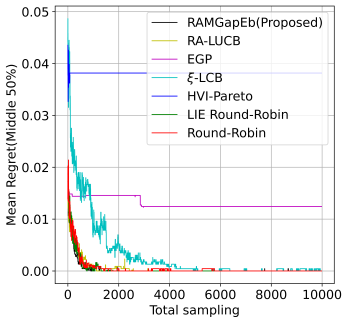}}
  \hfill
  \subfigure[Experiment 4 with $K = 100$ arms]{%
    \includegraphics[width=0.45\textwidth]{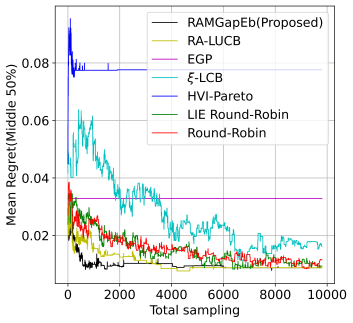}}
  \caption{\textbf{Comparison of average simple regret (middle 50\%) over total number of samples.} 
  }
  \label{fig:experiment3-4}
\end{figure}

\subsection{Results}

\subsubsection*{Evaluation under the Fixed-Confidence Setting}

We evaluated the performance of RAMGapE under the fixed-confidence setting by comparing 
with several baseline algorithms, including Round-Robin, Dominated Elimination Round-Robin (DE Round-Robin), and Risk-Averse LUCB (RA-LUCB).
As shown in Fig.~\ref{fig:experiment1}, in Experiment 1, the stopping times of RAMGapEc significantly shorter than those of Round-Robin and DE Round-Robin, and those of RAMGapEc and RA-LUCB are comparable with marginal differences (see also Fig.~\ref{fig:gaussian_stopping_time} in Appendix~\ref{sec:gaussian}). 
%
The visualization of confidence intervals at stopping time (Experiments 2.1 and 2.2, Figs.~\ref{fig:experiment2_1}-\ref{fig:experiment2_2}) further demonstrates that RAMGapEc, as well as RA-LUCB, effectively avoids unnecessary exploration of non-Pareto arms, focusing sampling efforts on arms near the Pareto frontier. 

\subsubsection*{Evaluation under the Fixed-Budget Setting}

In the fixed-budget setting, as shown in Fig.~\ref{fig:experiment3-4}, RAMGapE was compared with other existing approaches, including Round-Robin, LIE Round-Robin, RA-LUCB, $\xi$-LCB, EGP, and HVI-Pareto. The results of Experiments 3 ($K=10$) and 4 ($K=100$) show that RAMGapE exhibits a fast convergence of average simple regret both in small- and large-scale problems. Especially for the $K=100$ problem, RAMGapE exhibits the fastest drop in the mean regret with respect to the total sampling much faster than the comparable algorithm RA-LUCB in Experiments 1 and 2 in this experiment. Note also that, except RAMGapE, the other algorithms either converge very slowly or converge to some mean regrets larger than that acquired by RAMGapE. This suggests that our RAMGapE not only unifies the fixed budget and fixed confidence settings with different stopping criterion but also outperforms or equally best performs among the comparison algorithms for both settings.

Moreover, an analysis of the pulling ratios reveals how RAMGapE efficiently allocates samples. As shown in Figs.~\ref{fig:comp_select_e3} and~\ref{fig:comp_select_e4} (Appendix~\ref{subsec:appendix_pulling_ratios}), rather than naively focusing only on empirically optimal arms, RAMGapE strategically balances exploration between Pareto and non-Pareto arms to precisely identify the boundary of the optimal set. Once suboptimal arms are identified with sufficient confidence, the algorithm adaptively shifts its focus, resulting in a higher proportion of samples being allocated to promising, Pareto-optimal arms in later stages. This efficient exploration strategy contributes directly to its steady reduction of regret and the accurate identification of the Pareto set, as shown in Fig.~\ref{fig:experiment3-4}.

\subsubsection*{Overall Assessment}

Overall, RAMGapE demonstrated stable performance across both fixed-confidence and fixed-budget settings, efficiently balancing exploration and exploitation. The results highlight its suitability for risk-averse decision-making in stochastic environments where identifying multiple viable solutions is required.

\section{Conclusion}\label{sec:conclusion}

We presented RAMGapE (Risk-Averse Multi-objective Gap-based Exploration), a unified algorithmic framework for Risk-Averse Best Arm Set Identification problem that jointly optimizes both expected reward and risk via the \textit{mean-variance} (MV) criterion. Unlike conventional approaches that treat risk as an isolated objective, RAMGapE integrates risk directly into the multi-objective formulation, enabling principled identification of Pareto-optimal solutions that simultaneously balance utility and uncertainty.
%
We provided theoretical guarantees, including correctness and sample complexity bounds, and demonstrated 
that RAMGapE achieves efficient sampling and accurate identification of Pareto-optimal solutions. Our results show that RAMGapE adaptively concentrates sampling on uncertain regions near the Pareto frontier, while efficiently pruning non-Pareto arms far from Pareto fronts. 
This targeted exploration yields robust performance across both small- and large-scale problem instances.
A key strength of RAMGapE lies in its ability to flexibly allocate sampling resources toward high-uncertainty regions near the Pareto frontier, making it well-suited for real-world risk-sensitive applications such as medical trials or portfolio optimization, where both performance and risk must be jointly optimized.

Future work includes several directions. A key avenue is to extend the present framework to incorporate richer and more complex risk measures beyond mean-variance measure. For instance, adapting RAMGapE to spectral risk measures like CVaR or EVaR~\cite{ahmadi2012entropic} is a non-trivial but important challenge. This would require redefining the gap quantities in a way that accommodates these tail-focused risk measures and deriving new concentration bounds for their empirical estimators. The partial-order structure induced by such measures may also differ significantly from the one in the mean-variance space, demanding a careful redesign of the exploration strategy. Further theoretical challenges include addressing non-stationary environments and refining the analysis to derive tighter sample complexity bounds.
Overall, RAMGapE advances, with its unified formulation and significant performance, the state of risk-aware multi-objective bandit problem, providing a solid foundation for tackling complex, real-world decision-making problems under uncertainty.



\acks{The authors express their sincere gratitude to Professors Atsuyoshi Nakamura and Hiroshi Teramoto 
for their invaluable guidance and insightful discussions throughout this research.
This work was supported by JST/CREST Innovative Measurement and Analysis (Grant Number JPMJCR2333 to TK), JSPS Program for Forming Japan’s Peak
Research Universities (J-PEAKS) Grant Number JPJS00420230001, JSPS Grant in Aid for Scientific Research (A) (General) (Grant Number 24H00685 to KT) and JST SPRING (Grant Number JPMJSP2119 to SN).}


\bibliography{acml25}

\begin{thebibliography}{37}
\providecommand{\natexlab}[1]{#1}
\providecommand{\url}[1]{\texttt{#1}}
\expandafter\ifx\csname urlstyle\endcsname\relax
  \providecommand{\doi}[1]{doi: #1}\else
  \providecommand{\doi}{doi: \begingroup \urlstyle{rm}\Url}\fi

\bibitem[Ahmadi-Javid(2012)]{ahmadi2012entropic}
Amir Ahmadi-Javid.
\newblock Entropic value-at-risk: A new coherent risk measure.
\newblock \emph{Journal of Optimization Theory and Applications}, 155\penalty0 (3):\penalty0 1105--1123, 2012.

\bibitem[Audibert and Bubeck(2010)]{audibert2010best}
Jean-Yves Audibert and S{\'e}bastien Bubeck.
\newblock Best arm identification in multi-armed bandits.
\newblock In \emph{COLT-23th Conference on learning theory-2010}, pages 41--53, 2010.

\bibitem[Auer et~al.(2016)Auer, Chiang, Ortner, and Drugan]{auer2016pareto}
Peter Auer, Chao-Kai Chiang, Ronald Ortner, and Madalina Drugan.
\newblock Pareto front identification from stochastic bandit feedback.
\newblock In \emph{Artificial intelligence and statistics}, pages 939--947. PMLR, 2016.

\bibitem[Azuma(1967)]{azuma1967weighted}
Kazuoki Azuma.
\newblock Weighted sums of certain dependent random variables.
\newblock \emph{Tohoku Mathematical Journal, Second Series}, 19\penalty0 (3):\penalty0 357--367, 1967.

\bibitem[Bubeck et~al.(2013)Bubeck, Cesa-Bianchi, and Lugosi]{bubeck2013bandits}
S{\'e}bastien Bubeck, Nicolo Cesa-Bianchi, and G{\'a}bor Lugosi.
\newblock Bandits with heavy tail.
\newblock \emph{IEEE Transactions on Information Theory}, 59\penalty0 (11):\penalty0 7711--7717, 2013.

\bibitem[Cao et~al.(2015)Cao, Smucker, and Robinson]{cao2015using}
Yongtao Cao, Byran~J Smucker, and Timothy~J Robinson.
\newblock On using the hypervolume indicator to compare pareto fronts: Applications to multi-criteria optimal experimental design.
\newblock \emph{Journal of Statistical Planning and Inference}, 160:\penalty0 60--74, 2015.

\bibitem[Catoni(2012)]{catoni2012challenging}
Olivier Catoni.
\newblock Challenging the empirical mean and empirical variance: a deviation study.
\newblock In \emph{Annales de l'IHP Probabilit{\'e}s et statistiques}, volume~48, pages 1148--1185, 2012.

\bibitem[Chen et~al.(2022)Chen, Gangrade, and Saligrama]{chen2022strategies}
Tianrui Chen, Aditya Gangrade, and Venkatesh Saligrama.
\newblock Strategies for safe multi-armed bandits with logarithmic regret and risk.
\newblock In \emph{International Conference on Machine Learning}, pages 3123--3148. PMLR, 2022.

\bibitem[David and Shimkin(2016)]{david2016pure}
Yahel David and Nahum Shimkin.
\newblock Pure exploration for max-quantile bandits.
\newblock In \emph{Joint European conference on machine learning and knowledge discovery in databases}, pages 556--571. Springer, 2016.

\bibitem[Du et~al.(2021)Du, Wang, Fang, and Huang]{du2021continuous}
Yihan Du, Siwei Wang, Zhixuan Fang, and Longbo Huang.
\newblock Continuous mean-covariance bandits.
\newblock \emph{Advances in Neural Information Processing Systems}, 34:\penalty0 875--886, 2021.

\bibitem[Endris et~al.(2016)Endris, Stenzinger, Pfarr, Penzel, M{\"o}bs, Lenze, Darb-Esfahani, Hummel, Jung, Lehmann, et~al.]{ovariancancer}
Volker Endris, Albrecht Stenzinger, Nicole Pfarr, Roland Penzel, Markus M{\"o}bs, Dido Lenze, Silvia Darb-Esfahani, Michael Hummel, Andreas Jung, Ulrich Lehmann, et~al.
\newblock Ngs-based brca1/2 mutation testing of high-grade serous ovarian cancer tissue: results and conclusions of the first international round robin trial.
\newblock \emph{Virchows Archiv}, 468:\penalty0 697--705, 2016.

\bibitem[Even-Dar et~al.(2002)Even-Dar, Mannor, and Mansour]{even2002pac}
Eyal Even-Dar, Shie Mannor, and Yishay Mansour.
\newblock Pac bounds for multi-armed bandit and markov decision processes.
\newblock In \emph{Computational Learning Theory: 15th Annual Conference on Computational Learning Theory, COLT 2002 Sydney, Australia, July 8--10, 2002 Proceedings 15}, pages 255--270. Springer, 2002.

\bibitem[Gabillon et~al.(2012)Gabillon, Ghavamzadeh, and Lazaric]{gabillon2012best}
Victor Gabillon, Mohammad Ghavamzadeh, and Alessandro Lazaric.
\newblock Best arm identification: A unified approach to fixed budget and fixed confidence.
\newblock volume~25, 2012.

\bibitem[Honorio and Jaakkola(2014)]{honorio2014tight}
Jean Honorio and Tommi Jaakkola.
\newblock Tight bounds for the expected risk of linear classifiers and pac-bayes finite-sample guarantees.
\newblock In \emph{Artificial Intelligence and Statistics}, pages 384--392. PMLR, 2014.

\bibitem[Hou et~al.(2022)Hou, Tan, and Zhong]{hou2023almost}
Yunlong Hou, Vincent~YF Tan, and Zixin Zhong.
\newblock Almost optimal variance-constrained best arm identification.
\newblock \emph{IEEE Transactions on Information Theory}, 69\penalty0 (4):\penalty0 2603--2634, 2022.

\bibitem[Huo and Fu(2017)]{huo2017risk}
Xiaoguang Huo and Feng Fu.
\newblock Risk-aware multi-armed bandit problem with application to portfolio selection.
\newblock \emph{Royal Society open science}, 4\penalty0 (11):\penalty0 171377, 2017.

\bibitem[Kalyanakrishnan et~al.(2012)Kalyanakrishnan, Tewari, Auer, and Stone]{kalyanakrishnan2012pac}
Shivaram Kalyanakrishnan, Ambuj Tewari, Peter Auer, and Peter Stone.
\newblock Pac subset selection in stochastic multi-armed bandits.
\newblock In \emph{ICML}, volume~12, pages 655--662, 2012.

\bibitem[Karnin et~al.(2013)Karnin, Koren, and Somekh]{karnin2013almost}
Zohar Karnin, Tomer Koren, and Oren Somekh.
\newblock Almost optimal exploration in multi-armed bandits.
\newblock In \emph{International conference on machine learning}, pages 1238--1246. PMLR, 2013.

\bibitem[Keramati et~al.(2020)Keramati, Dann, Tamkin, and Brunskill]{keramati2020being}
Ramtin Keramati, Christoph Dann, Alex Tamkin, and Emma Brunskill.
\newblock Being optimistic to be conservative: Quickly learning a cvar policy.
\newblock In \emph{Proceedings of the AAAI conference on artificial intelligence}, volume~34, pages 4436--4443, 2020.

\bibitem[Kone et~al.(2023)Kone, Kaufmann, and Richert]{kone2023adaptive}
Cyrille Kone, Emilie Kaufmann, and Laura Richert.
\newblock Adaptive algorithms for relaxed pareto set identification.
\newblock \emph{Advances in Neural Information Processing Systems}, 36:\penalty0 35190--35201, 2023.

\bibitem[Kone et~al.(2024)Kone, Kaufmann, and Richert]{kone2024fixedbudget}
Cyrille Kone, Emilie Kaufmann, and Laura Richert.
\newblock Bandit pareto set identification: the fixed budget setting.
\newblock In \emph{International Conference on Artificial Intelligence and Statistics}, pages 2548--2556. PMLR, 2024.

\bibitem[Kone et~al.(2025)Kone, Kaufmann, and Richert]{kone2025bandit}
Cyrille Kone, Emilie Kaufmann, and Laura Richert.
\newblock Bandit pareto set identification in a multi-output linear model.
\newblock In \emph{Seventeenth European Workshop on Reinforcement Learning}, 2025.

\bibitem[Lattimore and Szepesv{\'a}ri(2020)]{lattimore2020bandit}
Tor Lattimore and Csaba Szepesv{\'a}ri.
\newblock \emph{Bandit algorithms}.
\newblock Cambridge University Press, Cambridge, United Kingdom, 2020.

\bibitem[Li et~al.(2010)Li, Chu, Langford, and Schapire]{li2010contextual}
Lihong Li, Wei Chu, John Langford, and Robert~E Schapire.
\newblock A contextual-bandit approach to personalized news article recommendation.
\newblock In \emph{Proceedings of the 19th international conference on World wide web}, pages 661--670, 2010.

\bibitem[Li et~al.(2011)Li, Chu, Langford, and Wang]{li2011unbiased}
Lihong Li, Wei Chu, John Langford, and Xuanhui Wang.
\newblock Unbiased offline evaluation of contextual-bandit-based news article recommendation algorithms.
\newblock In \emph{Proceedings of the fourth ACM international conference on Web search and data mining}, pages 297--306, 2011.

\bibitem[Madhukar et~al.(2017)Madhukar, Khade, Huang, Gayvert, Galletti, Stogniew, Allen, Giannakakou, and Elemento]{madhukar2017new}
Neel~S Madhukar, Prashant~K Khade, Linda Huang, Kaitlyn Gayvert, Giuseppe Galletti, Martin Stogniew, Joshua~E Allen, Paraskevi Giannakakou, and Olivier Elemento.
\newblock A new big-data paradigm for target identification and drug discovery.
\newblock \emph{Biorxiv}, page 134973, 2017.

\bibitem[Pannee et~al.(2016)Pannee, Gobom, Shaw, Korecka, Chambers, Lame, Jenkins, Mylott, Carrillo, Zegers, et~al.]{alzheimers}
Josef Pannee, Johan Gobom, Leslie~M Shaw, Magdalena Korecka, Erin~E Chambers, Mary Lame, Rand Jenkins, William Mylott, Maria~C Carrillo, Ingrid Zegers, et~al.
\newblock Round robin test on quantification of amyloid-$\beta$ 1--42 in cerebrospinal fluid by mass spectrometry.
\newblock \emph{Alzheimer's \& Dementia}, 12\penalty0 (1):\penalty0 55--59, 2016.

\bibitem[Qin et~al.(2014)Qin, Chen, and Zhu]{qin2014contextual}
Lijing Qin, Shouyuan Chen, and Xiaoyan Zhu.
\newblock Contextual combinatorial bandit and its application on diversified online recommendation.
\newblock In \emph{Proceedings of the 2014 SIAM International Conference on Data Mining}, pages 461--469. SIAM, 2014.

\bibitem[Rockafellar et~al.(2000)Rockafellar, Uryasev, et~al.]{rockafellar2000optimization}
R~Tyrrell Rockafellar, Stanislav Uryasev, et~al.
\newblock Optimization of conditional value-at-risk.
\newblock \emph{Journal of risk}, 2:\penalty0 21--42, 2000.

\bibitem[Sani et~al.(2012)Sani, Lazaric, and Munos]{sani2012risk}
Amir Sani, Alessandro Lazaric, and R{\'e}mi Munos.
\newblock Risk-aversion in multi-armed bandits.
\newblock \emph{Advances in neural information processing systems}, 25, 2012.

\bibitem[Shen et~al.(2022)Shen, Dunn, and Zavlanos]{shen2022risk}
Yi~Shen, Jessilyn Dunn, and Michael~M Zavlanos.
\newblock Risk-averse multi-armed bandits with unobserved confounders: A case study in emotion regulation in mobile health.
\newblock In \emph{2022 IEEE 61st Conference on Decision and Control (CDC)}, pages 144--149. IEEE, 2022.

\bibitem[Tamkin et~al.(2019)Tamkin, Keramati, Dann, and Brunskill]{tamkin2019distributionally}
Alex Tamkin, Ramtin Keramati, Christoph Dann, and Emma Brunskill.
\newblock Distributionally-aware exploration for cvar bandits.
\newblock In \emph{NeurIPS 2019 Workshop on Safety and Robustness on Decision Making}, 2019.

\bibitem[Tropp(2012)]{tropp2012user}
Joel~A Tropp.
\newblock User-friendly tail bounds for sums of random matrices.
\newblock \emph{Foundations of computational mathematics}, 12:\penalty0 389--434, 2012.

\bibitem[Ulrich et~al.(2008)Ulrich, Brockhoff, and Zitzler]{ulrich2008pattern}
Tamara Ulrich, Dimo Brockhoff, and Eckart Zitzler.
\newblock Pattern identification in pareto-set approximations.
\newblock In \emph{Proceedings of the 10th annual conference on Genetic and evolutionary computation}, pages 737--744, 2008.

\bibitem[Wainwright(2019)]{wainwright2019high}
Martin~J Wainwright.
\newblock \emph{High-dimensional statistics: A non-asymptotic viewpoint}, volume~48.
\newblock Cambridge university press, 2019.

\bibitem[Yang et~al.(2019)Yang, Emmerich, Deutz, and B{\"a}ck]{yang2019efficient}
Kaifeng Yang, Michael Emmerich, Andr{\'e} Deutz, and Thomas B{\"a}ck.
\newblock Efficient computation of expected hypervolume improvement using box decomposition algorithms.
\newblock \emph{Journal of Global Optimization}, 75:\penalty0 3--34, 2019.

\bibitem[Zitzler et~al.(2007)Zitzler, Brockhoff, and Thiele]{zitzler2007hypervolume}
Eckart Zitzler, Dimo Brockhoff, and Lothar Thiele.
\newblock The hypervolume indicator revisited: On the design of pareto-compliant indicators via weighted integration.
\newblock In \emph{Evolutionary Multi-Criterion Optimization: 4th International Conference, EMO 2007, Matsushima, Japan, March 5-8, 2007. Proceedings 4}, pages 862--876. Springer, 2007.

\end{thebibliography}

\clearpage

\appendix

\section{Geometric Interpretation of Gap Quantities}\label{illust_gap}

\begin{figure}[htb]
  \centering

  \subfigure[$M(k,i)< M(i,j)$]{%
    \includegraphics[width=0.48\textwidth]{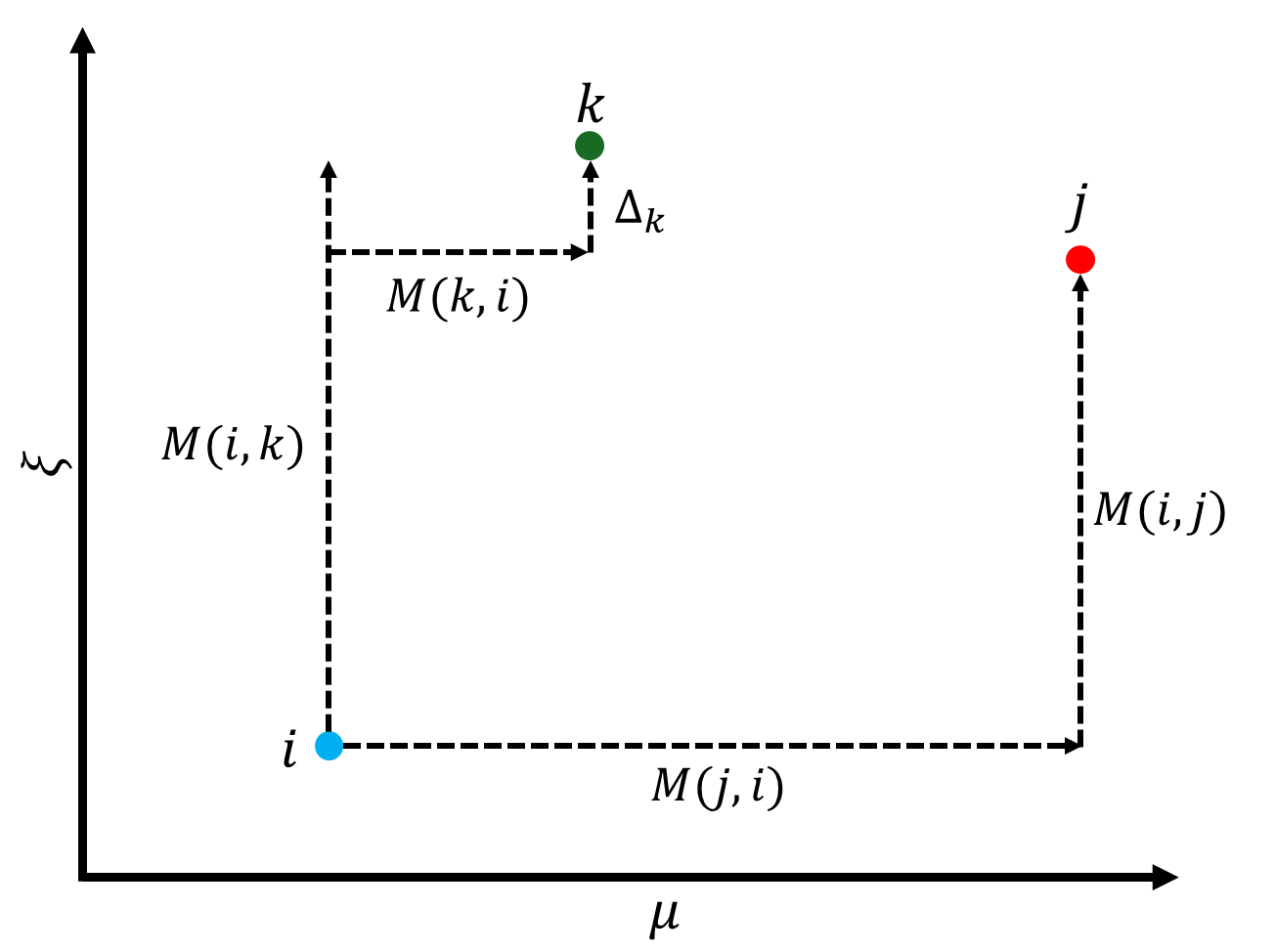}}
  \hfill
  \subfigure[$M(k,i)\geq M(i,j)$]{%
    \includegraphics[width=0.48\textwidth]{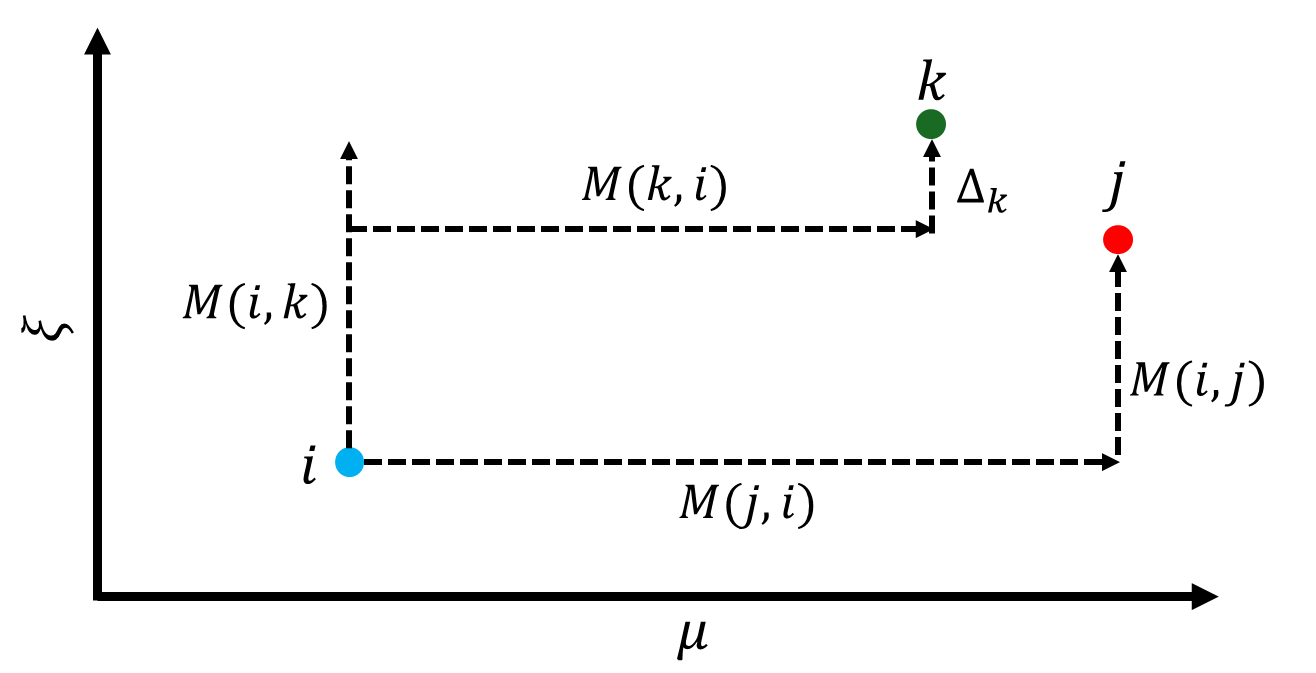}}
  \caption{\textbf{Geometric illustration of gap quantities in the mean-risk space.}
    Each point represents an arm, plotted by its expected reward on the horizontal axis ($\mu$) and scaled mean-variance risk on the vertical axis ($\xi := \alpha(\sigma^2 - \rho\mu)$). Here suppose that arms $i$ and $j$ are Pareto optimal while arm $k$ is non-Pareto, suboptimal. In both panels, arm $i$ has lower risk but a smaller mean than that of arm $k$; the opposite case (i.e., arm $i$ has a higher mean but a greater risk than those of arm $k$) can be treated similarly. If arm $i$ moves upward by more than $M(i,k)$ (or if $k$ moves downward), arm $i$ would become dominated by arm $k$. In these illustrations, the identity $M(i,k) = M(i,j) + (\xi_k - \xi_j)$ holds. Let us suppose that 
    the gap of arm $i$ satisfies $\Delta_i = \min \left( \min(M(i,j), M(j,i)), M(k,i)^+ + \Delta_k \right)$ under the existence of other possible Pareto and non-Pareto arms. Panel (a): When $M(k,i) < M(i,j)$, we obtain $M(i,k) = M(i,j) + (\xi_k - \xi_j) > M(k,i) + (\xi_k - \xi_j)=M^+(k,i) + \Delta_k$. The smaller $M(i,k)$ is, the more the samplings from arms $i$ and $k$ are required to discriminate them in practice. The choice of $M^+(k,i) + \Delta_k$---the lower bound of $M(i,k)$--- as the gap $\Delta_i$ corresponds to a ``conservative'' estimate reflecting 
    not only the suboptimal arm $k$ but also other Pareto arms $j$ via Eq.~(\ref{eq:def_gap}). Panel (b): When $M(k,i) \geq M(i,j)$, the term $\min(M(i,j), M(j,i))$ dominates the expression of $\Delta_i$, indicating that the difficulty in distinguishing $i$ from another Pareto-optimal arm $j$ outweighs that from suboptimal arm $k$. Thus, the contribution of $k$ to $\Delta_i$ becomes negligible in this case.
}
  \label{gap_illust}
\end{figure}

To clarify the role of the gap quantities introduced in Section \ref{sec2}, we provide a geometric illustration in the mean-risk space. Specifically, we consider the case in which a suboptimal arm $k \notin D^+$ is compared with multiple Pareto-optimal arms $i, j \in D^+$ (see Figure \ref{gap_illust}). The associated gap quantities characterize distinct sources of uncertainty that affect the accurate identification of $\epsilon$-Pareto optimal arms under sampling processes. 

\noindent
\textit{Robustness Against Elimination by Suboptimal Arms}: The quantity $M(k, i)^+ + \Delta_k$ represents a ``conservative'' margin ensuring that a Pareto-optimal arm $i \in D^+$ is not mistakenly eliminated due to statistical fluctuations in empirical estimates. Here, $M(k, i)^+$ measures how close the suboptimal arm $k$ is to dominating the optimal arm $i$, and $\Delta_k$ reflects the difficulty of confirming the suboptimality of $k$. Smaller value of this term indicates higher risk of erroneous elimination of $i$, by chance, along the process of some instance of samplings. 

\noindent
\textit{Distinguishability among Pareto-Optimal Arms}:
The term $\Delta_{ij} := \min\{ M(i, j), M(j, i) \}$ captures the closeness 
between two distinct Pareto-optimal arms $i, j \in D^+$. The smaller $\Delta_{ij}$, the closer the two arms in both objectives. This makes it more difficult to distinguish between them and judge correctly that neither of them dominates another with higher confidence. 

\noindent
\textit{Suboptimality Measure for Arm $k$}:
For a suboptimal arm $k \notin D^+$, the gap $\Delta_k$ represents the minimum shift required in all objectives for $k$ to enter the Pareto set. Smaller $\Delta_k$ implies that arm $k$ is closer to the Pareto frontier and, thus, more prone to be misclassified as Pareto-optimal, by chance. 


In summary, the gap quantity $\Delta_i$ for each arm $i \in [K]$ encodes the difficulty of correctly identifying the Pareto-optimal set under noisy and finite observations. It captures three key aspects: the resilience of optimal arms against being dominated by suboptimal ones, the distinguishability among optimal arms, and the closeness of suboptimal arms to the Pareto frontier. These interpretations provide a concrete understanding of the role and design of the gap-based arm selection strategy in our RAMGapE framework.

\section{Theoretical Analysis}\label{theoretical_analysis}

In this section, we present 
upper bounds on the performance of RAMGapEb and RAMGapEc, as introduced in Section \ref{sec3}. Our analysis follows a similar proof structure as the UGapE algorithm \cite{gabillon2012best}, which establishes a unified gap-based analysis framework for fixed-budget and fixed-confidence best arm identification. 
This similarity allows us to extend the classical regret arguments to the risk-averse multi-objective setting.  A key feature of RAMGapE is its unified arm selection strategy, which operates across both fixed-budget and fixed-confidence settings. This shared structure allows for a largely unified theoretical analysis. Appendix \ref{apd:first} outlines the common components of the proof, while Appendix \ref{CI_derivation} details the derivation of confidence intervals specific to the fixed-confidence setting. Before presenting the main theoretical results, we introduce the concept of an event $\mathcal{E}$ that will be essential for the following analysis.
\begin{equation}\label{event}
    \mathcal{E}\coloneq \Big\{\forall i\in[K],\forall t\in\{2K+1,\ldots,T\},|\hat{\mu}_i(t)-\mu_i|<\beta_i(t)\wedge\left|\hat{\mu}^{(2)}_i(t)-\mu^{(2)}_i\right|<\beta_i(t)\Big\},
\end{equation}
where the values of $T$ and $\beta_i(t)$ are defined separately for each setting. In particular, for any arm $i\in[K]$ and at any round $t \geq2K+1$, both $\underline{\mu}_i(t)\leq\mu_i\leq\overline{\mu}_i(t)$ and $\underline{\xi}_i(t)\leq\xi_i\leq\overline{\xi}_i(t)$ \textit{surely} hold when event $\mathcal{E}$ holds (see Appendix \ref{thm:fixed_conf}). 

\subsection{Analysis of the Arm Selection Strategy}\label{apd:first}

First, we present lower(Lemma \ref{lem:bound_v}) and upper(Lemma \ref{upper_bound_v}) for $V(t)$ in event $\mathcal{E}$, which indicates their connection with regret. 
We prove that for set $\widehat{D}^+_t~(\neq D^+)$ at any round $t\in\{2K+1,\ldots,T\}$, the quantity $V_{\widehat{D}^+_t \bigtriangleup D^+}(t)$ serves as an upper bound on the simple regret of this set $r_{\widehat{D}^+_t}$ under the condition that event $\mathcal{E}$ occurs.

\begin{lemma}\label{lem:bound_v}
    On event $\mathcal{E}$, 
    for any round $t\in\{2K+1,\ldots,T\}$, 
    we have $V(t)\geq r_{\widehat{D}^+_t}$.
\end{lemma}

\begin{proof}
On event $\mathcal{E}$, for any arm $i\in \widehat{D}^+_t\bigtriangleup D^+~(= D^+\setminus \widehat{D}^+_t \cup   \widehat{D}^+_t \setminus D^+)$ and each round $t\in\{2K+1,\ldots,T\}$, we prove the lemma for the two cases individually in parallel to the definition of gaps dependent on whether the arm in question belongs to the Pareto set $D^+$:

\noindent
\textbf{Case 1.}\ $i\in D^+\setminus \widehat{D}^+_t$: 

\noindent
\textbf{Case 1.1.}\ $\left|D^+\right|=1$: In this case, for any arm $k(\neq i)$, $i\succ k$ and ${M}(k,i)^+=0$. We can write
\begin{eqnarray}
    V_i(t)&=&\underset{j\in\widehat{D}^+_t\text{ s.t. }j\underset{t}{\succ}i}{\min}\max\Big(\overline{\mu}_i(t)-\underline{\mu}_j(t),\overline{\xi}_j(t)-\underline{\xi}_i(t)\Big)\nonumber\\
    &\geq&\underset{j\neq i}{\min}\max\Big(\overline{\mu}_i(t)-\underline{\mu}_j(t),\overline{\xi}_j(t)-\underline{\xi}_i(t)\Big)\nonumber\\
    &\stackrel{\mathrm{(A)}}{\geq}&
    \underset{j\neq i}{\min}\max({\mu}_i-{\mu}_j,{\xi}_j-{\xi}_i)\nonumber\\
    &\geq&
    \underset{j\neq i}{\min}\min({\mu}_i-{\mu}_j,{\xi}_j-{\xi}_i)\nonumber\\
    &=&\underset{j\notin D^+}{\min}\ m(j,i)\nonumber\\
    \label{Case1.1forlemma1}
    & \stackrel{\mathrm{(B)}}{=}&\underset{j\notin D^+}{\min}\ \Delta_j\nonumber\\
    &\stackrel{\mathrm{(C)}}{=}& \Delta_i= r_i(\widehat{D}^+_t)
\end{eqnarray}
The inequality (A) holds because of event $\mathcal{E}$, (B) holds from the definition of gap $\Delta_j$ (Eq.~\ref{eq:def_gap}) for $j \notin D^+$ where 
only a single Pareto solution exists, and (C) holds from that
for $i \in D^+$. 

\noindent
\textbf{Case 1.2.}\ $\left|D^+\right|\geq2$: In this case, we can write
\begin{eqnarray}
    V_i(t)&=&\underset{j\in\widehat{D}^+_t\text{ s.t. }j\underset{t}{\succ}i}{\min}\max\Big(\overline{\mu}_i(t)-\underline{\mu}_j(t),\overline{\xi}_j(t)-\underline{\xi}_i(t)\Big)\nonumber\\
    &\geq&\underset{j\neq i}{\min}\max\Big(\overline{\mu}_i(t)-\underline{\mu}_j(t),\overline{\xi}_j(t)-\underline{\xi}_i(t)\Big)\nonumber\\
    &\stackrel{\mathrm{(A)}}{\geq}&\underset{j\neq i}{\min}\max({\mu}_i-{\mu}_j,{\xi}_j-{\xi}_i)\nonumber\\
    &=&\min\left\{\underset{j \in D^+ \setminus \{i\}}{\min}\max({\mu}_i-{\mu}_j, {\xi}_j-{\xi}_i), \, \underset{j \notin D^+}{\min}\max({\mu}_i-{\mu}_j, {\xi}_j-{\xi}_i)\right\}\nonumber\\
    &=&\min\left\{\underset{j \in D^+ \setminus \{i\}}{\min} M(i, j), \, \underset{j \notin D^+}{\min} M(i, j)\right\}\nonumber\\
    &\geq&\min\left\{\underset{j \in D^+ \setminus \{i\}}{\min} \min \{M(i, j),M(j,i)\}, \, \underset{j \notin D^+}{\min} M(i, j)\right\}\nonumber\\
     &=&\min\left\{\underset{j \in D^+ \setminus \{i\}}{\min} \min \{M(i, j),M(j,i)\}, \,  M(i, k)\right\}\nonumber
\end{eqnarray}
where in the last equality we introduced
$k = \operatorname{argmin}_{j \notin D^+} M(i,j)$ for simplicity. The inequality (A) holds because of event $\mathcal{E}$.\\ The proof proceeds by considering two separate cases: $k \prec i$ (Case 1.2.1.) and $k \nprec i$ (Case 1.2.2.).

\noindent
\textbf{Case 1.2.1.} $k\prec i$: 

Suppose that arm $h$ satisfies $\Delta_k = \underset{j \in D^+\mathrm{~s.t.} j\succ k}{\operatorname{max}}\ {m}(k,j)=m(k,h)$ in defining the gap of the non-Pareto arm $k ~(\notin D^+)$.
Note that $M(i,k)\geq m(k,h)$ holds because otherwise it contradicts the condition of the arm $i$ being a Pareto solution: that is, suppose that $M(i,k)=\max(\mu_i-\mu_k,\xi_k-\xi_i)<\min(\mu_h-\mu_k,\xi_k-\xi_h)=m(k,h)$ holds. Then, when $M(i,k)=\mu_i-\mu_k$ is satisfied, $\mu_i-\mu_k<\min(\mu_h-\mu_k,\xi_k-\xi_h) \le \mu_h-\mu_k$, that is, $\mu_i < \mu_h$, apparently contradicts $i\in D^+$. This is the same for case $M(i,k)=\xi_k-\xi_i$.  Then, we can write
\begin{eqnarray}
    &&\min\left\{\underset{j \in D^+ \setminus \{i\}}{\min} \min \{M(i, j),M(j,i)\}, \,  M(i, k)\right\}\nonumber\\
    &\geq&\min\left\{\underset{j \in D^+ \setminus \{i\}}{\min} \min \{M(i, j),M(j,i)\}, \,  m(k, h)\right\}\nonumber\\
    &=&\min\left\{\underset{j \in D^+ \setminus \{i\}}{\min} \min \{M(i, j),M(j,i)\}, \,  \Delta_k\right\}\nonumber\\
    &=&\min\left\{\underset{j \in D^+ \setminus \{i\}}{\min} \min \{M(i, j),M(j,i)\}, \,  M(k,i)^++\Delta_k\right\} \quad (M(k,i)^+=0 \mathrm{~for~} \because k \prec i)\nonumber\\
    &\geq&
\min\left\{\underset{j \in D^+ \setminus \{i\}}{\min} \min\{M(i, j), M(j, i)\}, \, \underset{j \notin D^+}{\min} \left(M(j, i)^+ + \Delta_
    j \right)\right\}\nonumber\\
    \label{Case1.2.1forlemma1}&=&\Delta_i=r_i(\widehat{D}^+_t)
\end{eqnarray}
\textbf{Case 1.2.2.} $k\nprec i$: Because arm $k$ belongs not to the Pareto set, there should exist an arm $h\in D^+\setminus\{i\}$ such that $h\succ k$, that is, $\mu_h > \mu_k$ and $\xi_h < \xi_k$. Hence, 

\begin{align*}
    M(i, k) &= \max\{ \mu_i - \mu_k, \xi_k - \xi_i \} \\ 
    &\ge \max\{ \mu_i - \mu_h, \xi_h - \xi_i \} \\
    &= M(i, h) \\
    &\ge \underset{j \in D^+ \setminus \{i\}}{\min} \min \{M(i, j),M(j,i)\}.
\end{align*}
    
Therefore, 

\begin{eqnarray}
    &&\min\left\{\underset{j \in D^+ \setminus \{i\}}{\min} \min \{M(i, j),M(j,i)\}, \, M(i, k)\right\}\nonumber\\
    &=&\underset{j \in D^+ \setminus \{i\}}{\min} \min \{M(i, j),M(j,i)\}\nonumber\\
    &\geq&
\min\left\{\underset{j \in D^+ \setminus \{i\}}{\min} \min\{M(i, j), M(j, i)\}, \, \underset{j \notin D^+}{\min} \left(M(j, i)^+ + \Delta_
    j \right)\right\}\nonumber\\
    \label{Case1.2.2forlemma1}&=& r_i(\widehat{D}^+_t)
\end{eqnarray}
\textbf{Case 2.}\ $i\in \widehat{D}^+_t\setminus{D}^+$:
\begin{eqnarray}
    \label{Case2forlemma1}V_i(t)
    &=&\underset{j\neq i}{\max}\min\Big(\overline{\mu}_j(t)-\underline{\mu}_i(t),\overline{\xi}_i(t)-\underline{\xi}_j(t)\Big)\nonumber\\
    &\stackrel{\mathrm{(A)}}{=}&\underset{j\neq i}{\max}\min({\mu}_j-{\mu}_i,{\xi}_i-{\xi}_j)\nonumber\\
    &\geq&\underset{j\in D^+\text{ s.t. }j\succ i}{\max}\min({\mu}_j-{\mu}_i,{\xi}_i-{\xi}_j)\nonumber\\
    &=&\underset{j\in D^+\text{ s.t. }j\succ i}{\max}\ {m}(i,j)= \Delta_i = r_i(\widehat{D}^+_t)
\end{eqnarray}
The inequality (A) holds because of event $\mathcal{E}$.
\vskip\baselineskip
\noindent
Using Eq. \ref{Case1.1forlemma1}, \ref{Case1.2.1forlemma1}, \ref{Case1.2.2forlemma1} and \ref{Case2forlemma1}, we have
\begin{equation}
    V(t) \geq V_{\widehat{D}^+_t \bigtriangleup D^+}(t)=\underset{i\in \widehat{D}^+_t\bigtriangleup D^+}{\max}\ V_i(t)\geq\underset{i\in \widehat{D}^+_t\bigtriangleup D^+}{\max}\ r_i(\widehat{D}^+_t) \stackrel{\mathrm{(A)}}{=}\underset{i\in [K]}{\max}\ r_i(\widehat{D}^+_t)=r_{\widehat{D}^+_t},\nonumber
\end{equation}
where the equality (A) follows from that $r_i(\widehat{D}^+_t)=0$ 
for any $i\notin \widehat{D}^+_t\bigtriangleup D^+$. 
\end{proof}

\begin{lemma}\label{2beta}
On event $\mathcal{E}$, for any round $t\in\{2K+1,\ldots,T\}$, if arm $i\in\{m_t,p_t\}$ is pulled, we have $V(t)\leq 2\beta_i(t)$. 
\end{lemma}

\begin{proof}
The proof proceeds by case analysis, depending on whether $m_t\in\widehat{D}^+_t$ or not.
\vskip\baselineskip\noindent
\textbf{Case 1.} $m_t\in\widehat{D}_t^+$: In this case, we can write
\begin{eqnarray*}
    V(t)
    &=&
    \underset{j\neq m_t}{\max}\min\Big(\overline{\mu}_j(t)-\underline{\mu}_{m_t}(t),\overline{\xi}_{m_t}(t)-\underline{\xi}_j(t)\Big)\\
    &=&\min\Big(\overline{\mu}_{p_t}(t)-\underline{\mu}_{m_t}(t),\overline{\xi}_{m_t}(t)-\underline{\xi}_{p_t}(t)\Big)\\
    &=&
    \min\Big(\hat{\mu}_{p_t}(t)-\hat{\mu}_{m_t}(t),\hat{\xi}_{m_t}(t)-\hat{\xi}_{p_t}(t)\Big)+\beta_{m_t}(t)+\beta_{p_t}(t)\\
    &\leq&2\beta_i(t)
\end{eqnarray*}
\textbf{Case 2.}\ $m_t\notin\widehat{D}^+_t$: In this case, we can write
\begin{eqnarray*}
    V(t)&=&\underset{j\in\widehat{D}^+_t\text{ s.t. }j\underset{t}{\succ} m_t}{\min}\max\Big(\overline{\mu}_{m_t}(t)-\underline{\mu}_j(t),\overline{\xi}_j(t)-\underline{\xi}_{m_t}(t)\Big)\\
    &=&\max\Big(\overline{\mu}_{m_t}(t)-\underline{\mu}_{p_t}(t),\overline{\xi}_{p_t}(t)-\underline{\xi}_{m_t}(t)\Big)\\
    &=&\max\Big(\hat{\mu}_{m_t}(t)-\hat{\mu}_{p_t}(t),\hat{\xi}_{p_t}(t)-\hat{\xi}_{m_t}(t)\Big)+\beta_{m_t}(t)+\beta_{p_t}(t)\\
    &\leq&2\beta_i(t)
\end{eqnarray*}
The proof of Lemma \ref{2beta} is completed through the analysis of the two cases.
\end{proof}

\begin{lemma}\label{upper_bound_v}
On event $\mathcal{E}$, if arm $i\in\{m_t,p_t\}$ is pulled at time $t \in\{2K+1,\ldots,T\}$, we have
\begin{equation}
    V(t)\leq\min(0,-r_i(\widehat{D}^+_t)+2\beta_i(t))+2\beta_i(t).
\end{equation}    
\end{lemma}

\begin{proof}
    On event $\mathcal{E}$, for any round $t\in\{2K+1,\ldots,T\}$, if arm $i$ is pulled, from Lemma \ref{lem:bound_v} and Lemma \ref{2beta}, we have the inequalities:
\begin{equation*}
    r_i(\widehat{D}^+_t) \leq V(t) \leq 2 \beta_i(t).
\end{equation*}

Rearranging the left inequality yields:
\begin{equation*}
    0 \leq-r_i(\widehat{D}^+_t) +V(t)\leq -r_i(\widehat{D}^+_t) + 2 \beta_i(t)\Rightarrow0\leq -r_i(\widehat{D}^+_t) + 2 \beta_i(t).
\end{equation*}

Together with the right inequality,
\begin{equation*}
    V(t) \leq 2 \beta_i(t),
\end{equation*}
we combine these two inequalities to obtain
\begin{equation*}
    V(t) \leq \min\big(0, -r_i(\widehat{D}^+_t) + 2 \beta_i(t)\big) + 2 \beta_i(t).
\end{equation*}

This concludes the proof.
\end{proof}

\subsection{Regret Bound for the Fixed-Budget Setting}\label{theorem:fb_regret_bound}
Here we prove an upper-bound on the simple regret of RAMGapEb. Since the setting considered by the algorithm is fixed-budget, we may say $T=n$. From the definition of the confidence interval $\beta_i(t)$ in Eq. \ref{confidence_bound} and a union bound, we have that $\mathbb{P}[\mathcal{E}]\geq 1-4Kn\exp(-2a)$. We now have all the tools needed to prove the performance of RAMGapE for the $\epsilon$-Pareto set identification problem.
\setcounter{theorem}{1}
\begin{theorem}
If we run RAMGapEb with parameter $0<a\leq\frac{n-2K}{16K}\epsilon^2$, its simple regret $r_{\widehat{D}^+_n}$ satisfies
\begin{equation}
    \widetilde{\delta}=\mathbb{P}\left[r_{\widehat{D}_n}\geq\epsilon\right]\leq4Kn\exp(-2a),\nonumber
\end{equation}    
and in particular this probability is minimized for $a=\frac{n-2K}{16K}\epsilon^2$.
\end{theorem}
\begin{proof}
This proof is by contradiction. We assume that $r_{\widehat{D}^+_n}>\epsilon$ on event $\mathcal{E}$ and consider the following two steps:
\vskip\baselineskip\noindent
\textbf{Step 1}: Here we indicate that on the event $\mathcal{E}$, we have the following upper-bound on the number of pulls of any arm $i\in[K]$:
\begin{equation}\label{upper_bound_fb}
    T_i(n)<\frac{4a}{\max\left(\frac{r_i(\widehat{D}^+_n)+\epsilon}{2},\epsilon\right)^2}+2.
\end{equation}
Let $t_i$ be the last round that arm $i$ is pulled. If arm $i$ has been pulled only during the initialization phase, $T_i(n)=2$ and Eq. \ref{upper_bound_fb} trivially holds. If PullArm has selected $i$, then we have
\begin{equation}\label{connect_lemmas}
    \min(0,-r_i(\widehat{D}^+_{t_i})+2\beta_i(t_i))+2\beta_i(t_i)\stackrel{\mathrm{(A)}}{\geq} V(t_i)\stackrel{\mathrm{(B)}}{\geq} r_{
    \widehat{D}^+_{t_i}}>\epsilon.
\end{equation}
(A) and (B) hold because of Lemmas \ref{lem:bound_v} and \ref{upper_bound_v}.

We derive the following transformation by applying $\beta_i(t_i)$ and Eq. \ref{connect_lemmas}.
\begin{eqnarray}
    &&2\beta_i(t_i)\geq\max\Bigg(\frac{r_i(\widehat{D}^+_{t_i})+\epsilon}{2},\epsilon\Bigg)\Rightarrow4\beta^2_i(t_i)=\frac{4a}{T_i(t_i)}\geq\max\Bigg(\frac{r_i(\widehat{D}^+_{t_i})+\epsilon}{2},\epsilon\Bigg)^2\nonumber\\
    &\Leftrightarrow&T_i(t_i)\leq\frac{4a}{\max\left(\frac{r_i(\widehat{D}^+_{t_i})+\epsilon}{2},\epsilon\right)^2}<\frac{4a}{\max\left(\frac{r_i(\widehat{D}^+_{t_i})+\epsilon}{2},\epsilon\right)^2}+2\nonumber
\end{eqnarray}
As a result of the final transformation, we obtain Eq. \ref{upper_bound_fb}.
\vskip\baselineskip
\noindent
\textbf{Step 2}: 
Using Eq. \ref{upper_bound_fb}, we have $n = \sum_{i = 1}^K T_i(n) < \sum_{i=1}^{K}\frac{4a}{\max\left(\frac{r_i(\widehat{D}^+_n)+\epsilon}{2},\epsilon\right)^2}+2K$ on event $\mathcal{E}$. It is easy to see that by selecting $a\leq\frac{n-2K}{16K}\epsilon^2$, the right-hand-side of this inequality will be smaller than or equal to $n$, which is a contradiction. Thus, we conclude that $r_{\widehat{D}^+_n}\leq\epsilon$ on event $\mathcal{E}$. The final result follows from the probability of event $\mathcal{E}$ defined at the beginning of this section.
\end{proof}

\subsection{Regret Bound for Fixed-Confidence Setting}

We establish an upper bound on the simple regret of RAMGapEc. As the algorithm is analyzed in the fixed-confidence framework, we set $T = +\infty$ without loss of generality. By applying a union bound over all possible values of $T_i(t) \in \{2, \ldots, t\}$ for $t = 2K+1, \ldots, \infty$, and utilizing the confidence intervals $\beta_i(t)$ defined in Eq. \ref{confidence_bound}, it follows that the event $\mathcal{E}$ occurs with probability at least $1 - \delta$, i.e., $\mathbb{P}(\mathcal{E}) \ge 1 - \delta$ (see Theorem \ref{event_c}).

\begin{theorem}
    The RAMGapEc algorithm stops after $\widetilde{n}$ rounds and returns an $\epsilon$-Pareto set, $\widehat{D}^+_{\widetilde{n}}$, that satisfies
\begin{equation*}
    \mathbb{P}\left[r_{\widehat{D}^+_{\widetilde{n}}}\leq\epsilon\wedge\widetilde{n}\leq N\right]\geq1-\delta,
\end{equation*}
where $N=2K+ \mathcal{O}\left( \frac{K}{\epsilon^2} \log\left( \frac{K \log_2^2(1/\epsilon)}{\delta} \right) \right)$.
\end{theorem}

\begin{proof}
We first prove an upper bound on the simple regret of RAMGapEc. Using Lemma \ref{lem:bound_v}, we have that on the event $\mathcal{E}$, the simple regret of RAMGapEc upon stopping satisfies $V(t) \geq r_{\widehat{D}^+_{\widetilde{n}}}$. Since the algorithm stops when $V(t) < \epsilon$, this implies that $r_{\widehat{D}^+_{\widetilde{n}}} < \epsilon$ on $\mathcal{E}$, and hence
\begin{equation*}
\mathbb{P}\left[r_{\widehat{D}^+_{\widetilde{n}}} \leq \epsilon\right] \geq \mathbb{P}(\mathcal{E}) \geq 1 - \delta.
\end{equation*}

Next, we derive an upper bound on the number of times each arm is pulled. Let $t_i$ be the last round at which arm $i$ is selected. If arm $i$ is pulled only during the initialization phase, then $T_i(\widetilde{n}) = 2$ and the following bound holds trivially. We now consider the case where arm $i$ is selected at some round $t_i > 2K$ by the PullArm procedure. On the event $\mathcal{E}$, by Lemma \ref{2beta}, we have $V(t_i) \leq 2\beta_i(t_i)$. Combining this with the stopping condition $V(t_i) < \epsilon$, we obtain:
\begin{equation*}
\beta_i(t_i) < \frac{\epsilon}{2}.
\end{equation*}
Recall that the confidence interval is defined as
\begin{equation*}
\beta_i(t) = \sqrt{ \frac{4 \log\left( \frac{8K (\log_2 T_i(t))^2}{\delta} \right) }{T_i(t)} }.
\end{equation*}
Since RAMGapEc must hold for any arm $i$, $T_i(t_i)\geq\Omega(1/\epsilon^2)$ because of stopping criteria. Thus, it is natural to bound $\log_2T_i(t_i)\leq\log_2(1/\epsilon^2)$, and substitute accordingly.
\begin{equation*}
\log_2 T_i(t_i) \leq \log_2 \left( \frac{1}{\epsilon^2} \right) = 2\log_2(1/\epsilon),
\end{equation*}
and thus
\begin{equation*}
(\log_2 T_i(t_i))^2 \leq 4 \log_2^2(1/\epsilon).
\end{equation*}
Substituting this into the expression for $\beta_i(t_i)$ gives
\begin{equation*}
\beta_i(t_i) \leq \sqrt{ \frac{4 \log\left( \frac{32K \log_2^2(1/\epsilon)}{\delta} \right) }{T_i(t_i)} }.
\end{equation*}
To ensure $\beta_i(t_i) < \epsilon/2$, it suffices to require
\begin{equation*}
T_i(t_i) > \frac{16}{\epsilon^2} \log\left( \frac{32K \log_2^2(1/\epsilon)}{\delta} \right).
\end{equation*}
Hence, the number of pulls for any arm $i$ is upper-bounded as
\begin{equation*}\label{upper_bound_fc_sample}
    T_i(\widetilde{n}) \leq \frac{16}{\epsilon^2} \log\left( \frac{32K \log_2^2(1/\epsilon)}{\delta} \right) + 2.
\end{equation*}

Finally, summing over all arms $i \in [K]$, the total number of rounds before stopping satisfies
\begin{equation*}
\widetilde{n} = \sum_{i=1}^K T_i(\widetilde{n}) 
\leq \sum_{i=1}^K \left( \frac{16}{\epsilon^2} \log\left( \frac{32K \log_2^2(1/\epsilon)}{\delta} \right) + 2 \right)
= \mathcal{O}\left( \frac{K}{\epsilon^2} \log\left( \frac{K \log_2^2(1/\epsilon)}{\delta} \right) \right).
\end{equation*}
This completes the proof.
\end{proof}

\subsection{Other Theorems on RAMGapE}\label{apd:second}

\subsubsection{Output Accuracy}
\setcounter{theorem}{6}
\begin{theorem}\label{output}\label{thm:correctness}
Given allowance $\epsilon> 0$, on event $\mathcal{E}$, at any round $t \geq2K+1$, if $V(t)<\epsilon$, $\widehat{D}^+_t$ is $\epsilon$-Pareto set.
\end{theorem}
\begin{proof}
    The proof is completed by the following two conditions: 
\begin{eqnarray}
    \text{(1)}&& \text{if}\ i \in \widehat{D}^+_t\text{ and } V(t)<\epsilon,\ \text{then}\ \forall{j}\in[K], \neg(\mu_i\leq\mu_j-\epsilon\wedge\xi_i\geq\xi_j+\epsilon)\nonumber\\
    \text{(2)}&& \text{if}\ i \notin\widehat{D}^+_t\text{ and } V(t)<\epsilon,\ \text{then}\ \exists{j}\in[K]\setminus\{i\}, \mu_i\leq\mu_j+\epsilon\wedge\xi_i\geq\xi_j-\epsilon\nonumber
\end{eqnarray}
\noindent
Note that $V(t)\left(=\underset{k\in[K]}{\max}\ V_k(t)\right)<\epsilon$ implies $V_k(t) <\epsilon$ for any arm $k$.

\noindent
(1) Remind $V_i(t)=\underset{j\neq i}{\operatorname{max}}\ \min\Big(\overline{\mu}_j(t)-\underline{\mu}_i(t),\overline{\xi}_i(t)-\underline{\xi}_j(t)\Big) \text{ for arm }i\in \widehat{D}^+_t 
$ (Eq.~\ref{Vi(t)}). For arm $i\in\widehat{D}^+_t$, if $V_i(t) <\epsilon$, then $\overline{\mu}_k(t)-\underline{\mu}_i(t)<\epsilon\vee\overline{\xi}_i(t)-\underline{\xi}_k(t)<\epsilon$ holds for any arm $k ~(\neq i)$. On event $\mathcal{E}$ where the true mean and the true risk are surely within their confidence interval, $\mu_k-\mu_i<\epsilon\vee\xi_i-\xi_k<\epsilon$ holds. This implies (1). 
\vskip\baselineskip
\noindent
(2) $V_i(t)=\underset{j\in\widehat{D}^+_t \text{ s.t. }j\underset{t}{\succ} i}{\operatorname{min}}\ \max\Big(\overline{\mu}_i(t)-\underline{\mu}_j(t),\overline{\xi}_j(t)-\underline{\xi}_i(t)\Big)$ for arm $i\notin\widehat{D}^+_t$ (Eq.~\ref{Vi(t)}). For arm $i\notin\widehat{D}^+_t$, if $V_i(t) <\epsilon$, there exists an arm $k \underset{t}{\succ}i$ such that $\overline{\mu}_i(t)-\underline{\mu}_k(t)<\epsilon\wedge\overline{\xi}_k(t)-\underline{\xi}_i(t)<\epsilon$. On event $\mathcal{E}$, $\mu_i-\mu_k<\epsilon\wedge\xi_k-\xi_i<\epsilon$ holds. This implies (2). 

This completes the proof.

\end{proof}

\subsubsection{Derivation of high probability confidence interval under fixed confidence setting}\label{CI_derivation}
Here, we provide a proof of the theorem regarding the high-probability confidence intervals used in RAMGapEc.
\begin{theorem}\label{event_c}
    Under the fixed confidence setting, if let 
    $\beta_i(t)=\sqrt{4\log(8 K(\log_2 T_i(t))^2/\delta)/T_i(t)}$, 
    event $\mathcal{E}$ holds with probability at least $1-\delta$.
\end{theorem}\label{thm:fixed_conf}

\noindent
\begin{proof}
We define the events $\mathcal{E}_1$ and $\mathcal{E}_2$ as
\begin{eqnarray}
        \mathcal{E}_1
        &=&\Big\{\exists{i} \in [K], \exists{t} \geq2K+1, |\hat{\mu}_i(t)-\mu_i|\geq\beta_i(t) \Big\},\nonumber\\
        \mathcal{E}_2
        &=&\Big\{\exists{i} \in [K], \exists{t} \geq2K+1, |\hat{\mu}^{(2)}_i(t)-\mu^{(2)}_i|\geq\beta_i(t) \Big\}.\nonumber
\end{eqnarray}
From the definition of $\mathcal{E}$, the probability that $\mathcal{E}$ occurs is bounded from below by
$\mathbb{P}[\mathcal{E}]=1-\mathbb{P}\big[\mathcal{E}^\mathrm{C}\big]\geq1-(\mathbb{P}[\mathcal{E}_1]+\mathbb{P}[\mathcal{E}_2])$.

We derive 
upper bounds of each event $\mathcal{E}_1$ and $\mathcal{E}_2$. 

First, we provide an upper bound of the event $\mathcal{E}_1$.
%
%
From Hoeffding-Azuma inequality \cite{azuma1967weighted,tropp2012user}, for any integer $\gamma$ and a positive function $x(\gamma)$, the following inequality holds. 

\begin{equation}\label{HA_1}
    \mathbb{P}\Bigg[\exists i,\exists s\in\{1,\ldots,2^\gamma\},\left|\sum_{l=1}^{s} (X_i(l)-\mu_i) \right|>x(\gamma)\Bigg]
    \leq2\exp\left(-\frac{x(\gamma)^2}{2^\gamma}\right).
\end{equation}

Since the $\beta_i(t)$ depends only on $T_i(t)$ under given $K$ and $\delta$
we rewrite $\beta_i(t)$ as $\beta_{s}$ when $s = T_i(t)$, for convenience. 
Using Eq. \ref{HA_1}, $\mathbb{P}[\mathcal{E}_1]$ is bounded as follows:

\begin{eqnarray*}
    \mathbb{P}[\mathcal{E}_1] &\le& \sum_{i = 1}^K \mathbb{P}\left[\exists t \ge 2K+1,  |\hat{\mu}_i(t) - \mu_i|\ge \beta_i(t)\right] \\
    &\le& \sum_{i = 1}^K \mathbb{P}\left[\exists s \ge 2, \left|\sum_{l=1}^{s}(X_i(l)-\mu_i)\right|\ge s \beta_{s} \right]  \\
    &\le& \sum_{i = 1}^K \sum_{\gamma = 1}^{\infty} \mathbb{P}\Bigg[\exists s\in\{2^{\gamma-1},\ldots,2^\gamma\},\left|\sum_{l=1}^{s}(X_i(l)-\mu_i)\right| \ge 2^{\gamma-1} \beta_{2^{\gamma}} \Bigg] \\
    &\le& \sum_{i = 1}^K \sum_{\gamma = 1}^{\infty} \mathbb{P}\Bigg[\exists s\in\{1,\ldots,2^\gamma\},\left|\sum_{l=1}^{s}(X_i(l)-\mu_i)\right| \ge 2^{\gamma-1} \beta_{ 2^{\gamma}} \Bigg] \\
    &\le& \sum_{i = 1}^K \sum_{\gamma = 1}^{\infty} 2 \exp\left(- \frac{(2^{\gamma-1} \beta_{2^{\gamma}})^2}{2^\gamma}\right)\nonumber\\
    &=& \sum_{i = 1}^K \sum_{\gamma = 1}^{\infty} 2 \exp\left(- 2^{\gamma-2} \beta_{2^{\gamma}}^2\right)\nonumber\\
    &=& \sum_{i = 1}^K  \sum_{\gamma = 1}^{\infty} 2 \exp\left(- \log{\left(\frac{8 K(\log_2 2^\gamma)^2}{\delta}\right)} \right) \\
    &=& 2 K \sum_{\gamma = 1}^{\infty} \frac{\delta}{8 K \gamma^2} = \frac{\delta}{4} \frac{\pi^2}{6} < \frac{\delta}{2}
\end{eqnarray*}
In this study, we consider that arm $i$' reward distribution $\nu_i$ is bounded in $[0,1]$, so we can say for each $i\in[K]$, $X_i^{2}\in[0,1]$. So we can say $\mathbb{P}[\mathcal{E}_2]<2/\delta$ and conclude $\mathbb{P}[\mathcal{E}]\geq1-\delta$.
\end{proof}
\begin{corollary}
    As a consequence of Theorem \ref{event_c}, $\mathbb{P}\{\forall i, \forall t, |\widehat{\mathrm{MV}}_i(t)-{\mathrm{MV}}_i|<(3+\rho)\beta_i(t)\}\geq1-\delta$ holds.
\end{corollary}
\begin{proof}
    On event $\mathcal{E}$, we derive the following sequence of transformations:
\begin{eqnarray}
    (3+\rho)\beta_i(t)&=&\beta_i(t)+(2+\rho)\beta_i(t)\nonumber\\
    &>&|\hat{\mu}^{(2)}_i(t)-{\mu}^{(2)}_i|+(2+\rho)|\hat{\mu}_i(t)-\mu_i|\nonumber\\
    &=&|\hat{\mu}^{(2)}_i(t)-{\mu}^{(2)}_i|+2|\hat{\mu}_i(t)-\mu_i|+\rho|\hat{\mu}_i(t)-\mu_i|\nonumber\\
    &\geq&|\hat{\mu}^{(2)}_i(t)-{\mu}^{(2)}_i|+|\hat{\mu}_i(t)+\mu_i||\hat{\mu}_i(t)-\mu_i|+\rho|\hat{\mu}_i(t)-\mu_i|\nonumber\\
    &=&|\hat{\mu}^{(2)}_i(t)-{\mu}^{(2)}_i|+|\hat{\mu}^2_i(t)-\mu^2_i|+\rho|\hat{\mu}_i(t)-\mu_i|\nonumber\\
    &\geq&|\hat{\mu}^{(2)}_i(t)-{\mu}^{(2)}_i-\hat{\mu}^2_i(t)+\mu^{2}_i-\rho\hat{\mu}_i(t)+\rho\mu_i|\nonumber\\
    &=&|\hat{\mu}^{(2)}_i(t)-\hat{\mu}^{2}_i(t)-\rho\hat{\mu}_i(t)-\mu^{(2)}_i+\mu^2_i+\rho\mu_i|\nonumber\\
    &=&|\hat{\sigma}^2_i(t)-\rho\hat{\mu}_i(t)-\sigma^2_i+\rho\mu_i|=|\widehat{\mathrm{MV}}_i(t)-{\mathrm{MV}}_i|\nonumber
\end{eqnarray}
From the final result of the above derivation, it follows that under the event $\mathcal{E}$, the inequality $\forall i, \forall t, |\widehat{\mathrm{MV}}_i(t)-{\mathrm{MV}}_i|<(3+\rho)\beta_i(t)$ holds with probability at least $1 - \delta$.
\end{proof}

\subsubsection*{Analysis for Fixed Confidence Setting}

We now present the main theoretical guarantee of RAMGapEc in the fixed-confidence setting, showing that the returned solution is an $\varepsilon$-Pareto set with high probability.

We first prove that RAMGapEc terminates in finite time. 

\begin{theorem}[termination of RAMGapEc]
For the stopping time of  RAMGapEc $\tau := \inf \left\{t > 0 | V(t) < \epsilon \right\}$, 
\[
\mathbb{P}[\tau < \infty] = 1.
\]
\end{theorem}

\begin{proof}
In this proof, as in the proof of Theorem \ref{event_c},
we use the notation $\beta_n = \sqrt{4\log(8 K(\log_2 T_i(t))^2/\delta)/T_i(t)}$. 
At time round $t$, let $n$ be the smaller one of $T_{m_t}(t)$ and $T_{p_t}(t)$. Then, if $m_t \in \widehat{D}^+_t$, we have: 
\begin{align*}
\min\left\{\overline{\mu}_{p_t}(t) - \underline{\mu}_{m_t}(t), \; \overline{\xi}_{m_t}(t) - \underline{\xi}_{p_t} \right\} &\le \min\{\hat{\mu}_{p_t} - \hat{\mu}_{m_t}, \; \hat{\xi}_{m_t} - \hat{\xi}_{p_t}\} + \beta_{T_{m_t}(t)} + \beta_{T_{p_t}(t)} \\
    &\le 2 \beta_n.
\end{align*}

On the other hand, if $m_t \notin \widehat{D}^+_t$, then:
\begin{align*}
    \max\left\{\overline{\mu}_{m_t}(t) - \underline{\mu}_{p_t}(t), \; \overline{\xi}_{p_t}(t) - \underline{\xi}_{m_t} \right\} &\le \max\{\hat{\mu}_{p_t} - \hat{\mu}_{m_t}, \; \hat{\xi}_{m_t} - \hat{\xi}_{p_t}\} + \beta_{T_{m_t}(t)} + \beta_{T_{p_t}(t)} \\
    &\le 2 \beta_n.
\end{align*}

Therefore, we have $V(t) = V_{m_t}(t) \le 2 \beta_n$. 

Since $\beta_n \longrightarrow 0 $ as $n \longrightarrow \infty$, there exists some $N_0$ such that for all $n \ge N_0$, $V(t) < \epsilon$. 

Assume that RAMGapEc has not yest stopped at time $t$, that is, $V_{m_t}(t) > \epsilon$. 
Then, at least one of $T_{m_t}(t)$ or $T_{p_t}(t)$ is less than $N_0$. 
Since RAMGapEc selects the arm among $m_t$ and $p_t$ that has been pulled fewer times, the selected arm always has been pulled fewer than $N_0$ times. 
Therefore, RAMGapEc must stop by time round $t = K N_0$ at the latest. 
\end{proof}

\begin{theorem}[correctness of RAMGapEc]
Let $\varepsilon > 0$ and $\delta \in (0, 1)$ be user-specified accuracy and confidence parameters, respectively. Then, the output $\widehat{D}^+_{\tilde{n}}$ of RAMGapEc satisfies:
\[
\mathbb{P}\left[ \widehat{D}^+_{\tilde{n}} \text{ is an } \varepsilon\text{-Pareto set} \right] \geq 1 - \delta.
\]
\end{theorem}

\begin{proof}
Let $\mathcal{E}$ denote the high-probability event under which the empirical mean and second moment estimates are within the confidence intervals, as defined in Eq. \ref{event}. Theorem \ref{event_c} ensures that this event holds with probability at least $1 - \delta$, i.e.,
\[
\mathbb{P}[\mathcal{E}] \geq 1 - \delta.
\]

On event $\mathcal{E}$, the true values $\mu_i$ and $\xi_i$ of each arm $i \in [K]$ lie within the respective confidence bounds constructed by RAMGapEc at each round $t$. In particular, for all $t \geq 2K + 1$ and $i \in [K]$, we have:
\[
\mu_i \in [\hat{\mu}_i(t) \pm \beta_i(t)], \quad \xi_i \in [\hat{\xi}_i(t) \pm \beta_i(t)].
\]

From Theorem \ref{output}, we know that if $V(t) < \varepsilon$, then the current set $\widehat{D}^+_t$ is guaranteed to be an $\varepsilon$-Pareto set under event $\mathcal{E}$. RAMGapEc terminates at the first time $\tilde{n}$ such that $V(\tilde{n}) < \varepsilon$, and thus returns $\widehat{D}^+_{\tilde{n}}$.

Therefore, on event $\mathcal{E}$, the returned set $\widehat{D}^+_{\tilde{n}}$ satisfies the $\varepsilon$-Pareto optimality condition. Combining this with the probability bound on $\mathcal{E}$, we conclude:
\[
\mathbb{P}\left[ \widehat{D}^+_{\tilde{n}} \text{ is an } \varepsilon\text{-Pareto set} \right] \geq \mathbb{P}[\mathcal{E}] \geq 1 - \delta.
\]
\end{proof}

\section{Arm Setting for Experiments}

In this section, we describe the parameters of the arms used in the experiments.

\begin{table}[h]
\centering
\normalsize
\caption{$(a, b)$ values of each pattern for Experiment 3}\label{table_10_pattarns_b}
\begin{tabular}{|c|c|c|c|}
\hline
Index & $(a,b)$ & Index & $(a,b)$ \\ \hline
1  & (0.6462, 0.4308)  & 2 & (0.9146, 0.6684)  \\ \hline
3  & (0.3139, 0.2511)  & 4 & (3.7333, 3.2667)  \\ \hline
5  & (0.2028, 0.1940)  & 6 & (0.4050, 0.4234)  \\ \hline
7  & (1.1172, 1.2768)  & 8 & (1.6569, 2.0712)  \\ \hline
9  & (9.8779, 13.5171)  & 10 & (0.0800, 0.1200)  \\ \hline
\end{tabular}
\end{table}

\begin{table}[tb]
\centering
\normalsize
\caption{$(a, b)$ values of each pattern for Experiment 4}\label{table_100_patterns_b}\label{table_100_pattarns_b}
\begin{tabular}{|c|c|c|c|c|c|}
\hline
Index & $(a,b)$ & Index & $(a,b)$ & Index & $(a,b)$ \\ \hline
1 & (3.1125, 2.0750) & 2 & (0.3721, 0.2502) & 3 & (0.7343, 0.4978) \\ \hline
4 & (2.7665, 1.8914) & 5 & (1.5858, 1.0933) & 6 & (0.1750, 0.1217) \\ \hline
7 & (0.4844, 0.3396) & 8 & (7.4555, 5.2703) & 9 & (0.5855, 0.4173) \\ \hline
10 & (0.2011, 0.1445) & 11 & (1.0850, 0.7864) & 12 & (0.1700, 0.1242) \\ \hline
13 & (1.0451, 0.7700) & 14 & (0.8003, 0.5946) & 15 & (0.4112, 0.3080) \\ \hline
16 & (0.9397, 0.7098) & 17 & (0.1496, 0.1139) & 18 & (0.7701, 0.5913) \\ \hline
19 & (0.1644, 0.1273) & 20 & (0.1433, 0.1118) & 21 & (0.7420, 0.5839) \\ \hline
22 & (4.1542, 3.2963) & 23 & (0.2368, 0.1894) & 24 & (8.1278, 6.5557) \\ \hline
25 & (0.9541, 0.7758) & 26 & (0.8345, 0.6842) & 27 & (1.3518, 1.1174) \\ \hline
28 & (0.5351, 0.4459) & 29 & (1.3968, 1.1735) & 30 & (0.2526, 0.2140) \\ \hline
31 & (0.4995, 0.4266) & 32 & (0.1937, 0.1668) & 33 & (0.2171, 0.1885) \\ \hline
34 & (10.6034, 9.2780) & 35 & (0.2873, 0.2535) & 36 & (0.5752, 0.5115) \\ \hline
37 & (2.0854, 1.8697) & 38 & (0.5376, 0.4859) & 39 & (0.4057, 0.3697) \\ \hline
40 & (0.4173, 0.3834) & 41 & (0.7743, 0.7170) & 42 & (1.1156, 1.0416) \\ \hline
43 & (0.1817, 0.1710) & 44 & (0.3196, 0.3033) & 45 & (0.6128, 0.5861) \\ \hline
46 & (0.3385, 0.3265) & 47 & (0.6274, 0.6099) & 48 & (2.5964, 2.5444) \\ \hline
49 & (4.8632, 4.8046) & 50 & (1.2070, 1.2021) & 51 & (2.3015, 2.3108) \\ \hline
52 & (1.0011, 1.0134) & 53 & (0.1484, 0.1514) & 54 & (0.5733, 0.5897) \\ \hline
55 & (4.3481, 4.5092) & 56 & (5.7451, 6.0063) & 57 & (2.1309, 2.2458) \\ \hline
58 & (3.1815, 3.3804) & 59 & (0.2459, 0.2633) & 60 & (5.0978, 5.5048) \\ \hline
61 & (11.4694, 12.4856) & 62 & (1.3892, 1.5246) & 63 & (0.5330, 0.5897) \\ \hline
64 & (2.8987, 3.2332) & 65 & (1.7749, 1.9959) & 66 & (1.4193, 1.6090) \\ \hline
67 & (0.4206, 0.4807) & 68 & (0.7348, 0.8466) & 69 & (1.2393, 1.4396) \\ \hline
70 & (0.2078, 0.2433) & 71 & (0.2217, 0.2617) & 72 & (0.8207, 0.9769) \\ \hline
73 & (0.7387, 0.8865) & 74 & (0.1879, 0.2274) & 75 & (0.3771, 0.4600) \\ \hline
76 & (0.5618, 0.6909) & 77 & (0.2282, 0.2830) & 78 & (0.1043, 0.1303) \\ \hline
79 & (0.2410, 0.3037) & 80 & (1.8028, 2.2908) & 81 & (3.5191, 4.5084) \\ \hline
82 & (0.4082, 0.5273) & 83 & (0.1622, 0.2112) & 84 & (0.2656, 0.3488) \\ \hline
85 & (0.1022, 0.1354) & 86 & (1.5086, 2.0139) & 87 & (0.3365, 0.4530) \\ \hline
88 & (1.3058, 1.7721) & 89 & (0.2386, 0.3266) & 90 & (0.2890, 0.3988) \\ \hline
91 & (0.9339, 1.2993) & 92 & (0.8941, 1.2543) & 93 & (0.2227, 0.3151) \\ \hline
94 & (0.8204, 1.1703) & 95 & (1.3010, 1.8714) & 96 & (0.2659, 0.3856) \\ \hline
97 & (2.2502, 3.2913) & 98 & (2.7231, 4.0166) & 99 & (1.3219, 1.9663) \\ \hline
100 & (6.5372, 9.8058) &  &  &  &  \\ \hline
\end{tabular}
\end{table}

\begin{landscape}  
\begin{longtable}{c||c c c c c c c c c c c}
    \caption{$(a, b)$ values of each pattern for Experiments 1 and 2} \label{table_50_pattarns}\\
    \toprule
    pattern & \multicolumn{10}{c}{Index $(a, b)$}  \\
    \midrule
    \endfirsthead

    \toprule
    pattern & \multicolumn{10}{c}{Index $(a, b)$} \\
    \midrule
    \endhead

    \midrule
    \multicolumn{11}{r}{Go to next page} \\
    \midrule
    \endfoot

    \bottomrule
    \endlastfoot
    \midrule
    1  & 1  & (2.1574, 1.4383) & 2  & (0.2101, 0.1536) & 3  & (0.8969, 0.7175) & 4  & (0.1304, 0.1141) & 5  & (3.5940, 3.4377) \\
& 6  & (0.4050, 0.4234) & 7  & (0.2695, 0.3080) & 8  & (1.0520, 1.3150) & 9  & (9.8779, 13.5171) & 10  & (0.4308, 0.6462) \\
    \midrule
    2  & 1  & (0.9247, 0.6165) & 2  & (0.6420, 0.4691) & 3  & (0.3139, 0.2511) & 4  & (3.7333, 3.2667) & 5  & (1.9345, 1.8504) \\
& 6  & (0.1219, 0.1275) & 7  & (0.1826, 0.2087) & 8  & (1.0520, 1.3150) & 9  & (9.8779, 13.5171) & 10  & (0.3024, 0.4537) \\
    \midrule
    3  & 1  & (2.1574, 1.4383) & 2  & (0.6420, 0.4691) & 3  & (0.1303, 0.1043) & 4  & (0.8722, 0.7631) & 5  & (1.2305, 1.1770) \\
& 6  & (0.2854, 0.2983) & 7  & (3.2667, 3.7333) & 8  & (0.1690, 0.2113) & 9  & (0.3314, 0.4536) & 10  & (9.2000, 13.8000) \\
    \midrule
    4  & 1  & (0.3127, 0.2085) & 2  & (1.3443, 0.9823) & 3  & (0.6315, 0.5052) & 4  & (0.1304, 0.1141) & 5  & (12.2604, 11.7273) \\
& 6  & (3.4377, 3.5940) & 7  & (0.1826, 0.2087) & 8  & (1.6569, 2.0712) & 9  & (0.6684, 0.9146) & 10  & (0.3024, 0.4537) \\
    \midrule
    5  & 1  & (0.1200, 0.0800) & 2  & (0.9146, 0.6684) & 3  & (13.1619, 10.5295) & 4  & (0.4379, 0.3832) & 5  & (0.5941, 0.5683) \\
& 6  & (0.1940, 0.2028) & 7  & (0.2695, 0.3080) & 8  & (1.0520, 1.3150) & 9  & (2.8885, 3.9527) & 10  & (1.4383, 2.1574) \\
    \midrule
    6  & 1  & (2.1574, 1.4383) & 2  & (0.2101, 0.1536) & 3  & (13.1619, 10.5295) & 4  & (0.6154, 0.5385) & 5  & (1.2305, 1.1770) \\
& 6  & (0.8046, 0.8412) & 7  & (0.1141, 0.1304) & 8  & (3.0829, 3.8536) & 9  & (0.2306, 0.3156) & 10  & (0.3024, 0.4537) \\
    \midrule
    7  & 1  & (0.3127, 0.2085) & 2  & (0.2101, 0.1536) & 3  & (3.8536, 3.0829) & 4  & (2.0085, 1.7574) & 5  & (0.1275, 0.1219) \\
& 6  & (0.8046, 0.8412) & 7  & (11.1481, 12.7407) & 8  & (1.0520, 1.3150) & 9  & (0.4691, 0.6420) & 10  & (0.3024, 0.4537) \\
    \midrule
    8  & 1  & (0.6462, 0.4308) & 2  & (0.9146, 0.6684) & 3  & (0.3139, 0.2511) & 4  & (3.7333, 3.2667) & 5  & (0.2028, 0.1940) \\
& 6  & (0.4050, 0.4234) & 7  & (1.1172, 1.2768) & 8  & (1.6569, 2.0712) & 9  & (9.8779, 13.5171) & 10  & (0.0800, 0.1200) \\
    \midrule
    9  & 1  & (13.8000, 9.2000) & 2  & (1.3443, 0.9823) & 3  & (0.1303, 0.1043) & 4  & (0.6154, 0.5385) & 5  & (0.2983, 0.2854) \\
& 6  & (0.4050, 0.4234) & 7  & (1.7574, 2.0085) & 8  & (0.7175, 0.8969) & 9  & (0.1536, 0.2101) & 10  & (2.6857, 4.0286) \\
    \midrule
    10 & 1  & (0.2050, 0.1366) & 2  & (0.3156, 0.2306) & 3  & (0.8969, 0.7175) & 4  & (0.1304, 0.1141) & 5  & (1.2305, 1.1770) \\
& 6  & (3.4377, 3.5940) & 7  & (0.5385, 0.6154) & 8  & (10.5295, 13.1619) & 9  & (1.5501, 2.1213) & 10  & (0.3024, 0.4537) \\
    \midrule
    11 & 1  & (2.1574, 1.4383) & 2  & (0.9146, 0.6684) & 3  & (3.8536, 3.0829) & 4  & (0.6154, 0.5385) & 5  & (0.1275, 0.1219) \\
& 6  & (0.2854, 0.2983) & 7  & (0.1826, 0.2087) & 8  & (10.5295, 13.1619) & 9  & (0.9823, 1.3443) & 10  & (0.3024, 0.4537) \\
    \midrule
    12 & 1  & (13.8000, 9.2000) & 2  & (1.3443, 0.9823) & 3  & (3.8536, 3.0829) & 4  & (0.2087, 0.1826) & 5  & (0.1275, 0.1219) \\
& 6  & (0.2854, 0.2983) & 7  & (0.7631, 0.8722) & 8  & (0.5052, 0.6315) & 9  & (1.5501, 2.1213) & 10  & (0.3024, 0.4537) \\
    \midrule
    13 & 1  & (0.3127, 0.2085) & 2  & (2.1213, 1.5501) & 3  & (0.6315, 0.5052) & 4  & (12.7407, 11.1481) & 5  & (0.1275, 0.1219) \\
& 6  & (0.4050, 0.4234) & 7  & (0.1826, 0.2087) & 8  & (3.0829, 3.8536) & 9  & (0.9823, 1.3443) & 10  & (0.6165, 0.9247) \\
    \midrule
    14 & 1  & (13.8000, 9.2000) & 2  & (0.4536, 0.3314) & 3  & (0.2113, 0.1690) & 4  & (2.0085, 1.7574) & 5  & (0.1275, 0.1219) \\
& 6  & (3.4377, 3.5940) & 7  & (1.1172, 1.2768) & 8  & (0.7175, 0.8969) & 9  & (0.2306, 0.3156) & 10  & (0.4308, 0.6462) \\
    \midrule
    15 & 1  & (2.1574, 1.4383) & 2  & (0.6420, 0.4691) & 3  & (13.1619, 10.5295) & 4  & (3.7333, 3.2667) & 5  & (0.1275, 0.1219) \\
& 6  & (0.1940, 0.2028) & 7  & (0.3832, 0.4379) & 8  & (0.7175, 0.8969) & 9  & (0.9823, 1.3443) & 10  & (0.2085, 0.3127) \\
    \midrule
    16 & 1  & (13.8000, 9.2000) & 2  & (0.4536, 0.3314) & 3  & (0.6315, 0.5052) & 4  & (3.7333, 3.2667) & 5  & (0.8412, 0.8046) \\
& 6  & (1.8504, 1.9345) & 7  & (0.2695, 0.3080) & 8  & (0.1043, 0.1303) & 9  & (0.1536, 0.2101) & 10  & (0.9091, 1.3636) \\
    \midrule
    17 & 1  & (0.3127, 0.2085) & 2  & (0.6420, 0.4691) & 3  & (13.1619, 10.5295) & 4  & (0.1304, 0.1141) & 5  & (0.2028, 0.1940) \\
& 6  & (0.8046, 0.8412) & 7  & (3.2667, 3.7333) & 8  & (0.3585, 0.4482) & 9  & (1.5501, 2.1213) & 10  & (0.9091, 1.3636) \\
    \midrule
    18 & 1  & (0.4537, 0.3024) & 2  & (13.5171, 9.8779) & 3  & (0.8969, 0.7175) & 4  & (0.1304, 0.1141) & 5  & (0.2028, 0.1940) \\
& 6  & (0.2854, 0.2983) & 7  & (1.1172, 1.2768) & 8  & (1.6569, 2.0712) & 9  & (2.8885, 3.9527) & 10  & (0.4308, 0.6462) \\
    \midrule
    19 & 1  & (2.1574, 1.4383) & 2  & (0.6420, 0.4691) & 3  & (0.4482, 0.3585) & 4  & (0.2087, 0.1826) & 5  & (0.8412, 0.8046) \\
& 6  & (11.7273, 12.2604) & 7  & (1.1172, 1.2768) & 8  & (0.2511, 0.3139) & 9  & (2.8885, 3.9527) & 10  & (0.0800, 0.1200) \\
    \midrule
    20 & 1  & (0.6462, 0.4308) & 2  & (0.3156, 0.2306) & 3  & (0.1303, 0.1043) & 4  & (1.2768, 1.1172) & 5  & (3.5940, 3.4377) \\
& 6  & (0.8046, 0.8412) & 7  & (0.1826, 0.2087) & 8  & (1.6569, 2.0712) & 9  & (0.3314, 0.4536) & 10  & (9.2000, 13.8000) \\
    \midrule
    21 & 1  & (2.1574, 1.4383) & 2  & (0.3156, 0.2306) & 3  & (0.4482, 0.3585) & 4  & (0.8722, 0.7631) & 5  & (0.5941, 0.5683) \\
& 6  & (0.1219, 0.1275) & 7  & (0.1826, 0.2087) & 8  & (10.5295, 13.1619) & 9  & (2.8885, 3.9527) & 10  & (0.9091, 1.3636) \\
    \midrule
    22 & 1  & (0.3127, 0.2085) & 2  & (0.4536, 0.3314) & 3  & (13.1619, 10.5295) & 4  & (0.8722, 0.7631) & 5  & (0.1275, 0.1219) \\
& 6  & (0.1940, 0.2028) & 7  & (1.7574, 2.0085) & 8  & (0.5052, 0.6315) & 9  & (2.8885, 3.9527) & 10  & (0.9091, 1.3636) \\
    \midrule
    23 & 1  & (13.8000, 9.2000) & 2  & (0.4536, 0.3314) & 3  & (1.3150, 1.0520) & 4  & (2.0085, 1.7574) & 5  & (0.8412, 0.8046) \\
& 6  & (0.5683, 0.5941) & 7  & (0.2695, 0.3080) & 8  & (3.0829, 3.8536) & 9  & (0.0928, 0.1270) & 10  & (0.1366, 0.2050) \\
    \midrule
    24 & 1  & (0.6462, 0.4308) & 2  & (0.1270, 0.0928) & 3  & (1.3150, 1.0520) & 4  & (0.2087, 0.1826) & 5  & (0.8412, 0.8046) \\
& 6  & (3.4377, 3.5940) & 7  & (0.3832, 0.4379) & 8  & (10.5295, 13.1619) & 9  & (0.2306, 0.3156) & 10  & (1.4383, 2.1574) \\
    \midrule
    25 & 1  & (2.1574, 1.4383) & 2  & (0.9146, 0.6684) & 3  & (0.3139, 0.2511) & 4  & (12.7407, 11.1481) & 5  & (3.5940, 3.4377) \\
& 6  & (0.4050, 0.4234) & 7  & (0.1141, 0.1304) & 8  & (0.5052, 0.6315) & 9  & (0.1536, 0.2101) & 10  & (0.9091, 1.3636) \\
    \midrule
    26& 1  & (0.4537, 0.3024) & 2  & (0.9146, 0.6684) & 3  & (13.1619, 10.5295) & 4  & (0.2087, 0.1826) & 5  & (1.9345, 1.8504) \\
& 6  & (0.2854, 0.2983) & 7  & (1.1172, 1.2768) & 8  & (0.5052, 0.6315) & 9  & (0.0928, 0.1270) & 10  & (2.6857, 4.0286) \\
    \midrule
    27 & 1  & (0.1200, 0.0800) & 2  & (2.1213, 1.5501) & 3  & (3.8536, 3.0829) & 4  & (0.6154, 0.5385) & 5  & (12.2604, 11.7273) \\
& 6  & (0.1940, 0.2028) & 7  & (0.7631, 0.8722) & 8  & (0.3585, 0.4482) & 9  & (0.9823, 1.3443) & 10  & (0.2085, 0.3127) \\
    \midrule
    28 & 1  & (13.8000, 9.2000) & 2  & (3.9527, 2.8885) & 3  & (0.3139, 0.2511) & 4  & (2.0085, 1.7574) & 5  & (0.2028, 0.1940) \\
& 6  & (0.8046, 0.8412) & 7  & (0.3832, 0.4379) & 8  & (0.1043, 0.1303) & 9  & (0.9823, 1.3443) & 10  & (0.4308, 0.6462) \\
    \midrule
    29 & 1  & (0.6462, 0.4308) & 2  & (0.3156, 0.2306) & 3  & (1.3150, 1.0520) & 4  & (0.1304, 0.1141) & 5  & (1.9345, 1.8504) \\
& 6  & (11.7273, 12.2604) & 7  & (0.1826, 0.2087) & 8  & (0.3585, 0.4482) & 9  & (0.6684, 0.9146) & 10  & (2.6857, 4.0286) \\
    \midrule
    30 & 1  & (0.9247, 0.6165) & 2  & (0.2101, 0.1536) & 3  & (0.3139, 0.2511) & 4  & (0.4379, 0.3832) & 5  & (0.5941, 0.5683) \\
& 6  & (1.1770, 1.2305) & 7  & (3.2667, 3.7333) & 8  & (1.6569, 2.0712) & 9  & (0.0928, 0.1270) & 10  & (9.2000, 13.8000) \\
    \midrule
    31 & 1  & (0.9247, 0.6165) & 2  & (0.2101, 0.1536) & 3  & (0.3139, 0.2511) & 4  & (0.4379, 0.3832) & 5  & (3.5940, 3.4377) \\
& 6  & (0.5683, 0.5941) & 7  & (1.7574, 2.0085) & 8  & (0.1043, 0.1303) & 9  & (9.8779, 13.5171) & 10  & (0.9091, 1.3636) \\
    \midrule
    32 & 1  & (4.0286, 2.6857) & 2  & (0.2101, 0.1536) & 3  & (0.4482, 0.3585) & 4  & (12.7407, 11.1481) & 5  & (0.5941, 0.5683) \\
& 6  & (1.1770, 1.2305) & 7  & (1.7574, 2.0085) & 8  & (0.7175, 0.8969) & 9  & (0.0928, 0.1270) & 10  & (0.2085, 0.3127) \\
    \midrule
    33 & 1  & (0.3127, 0.2085) & 2  & (0.2101, 0.1536) & 3  & (0.4482, 0.3585) & 4  & (0.1304, 0.1141) & 5  & (12.2604, 11.7273) \\
& 6  & (3.4377, 3.5940) & 7  & (0.5385, 0.6154) & 8  & (1.6569, 2.0712) & 9  & (0.9823, 1.3443) & 10  & (0.6165, 0.9247) \\
    \midrule
    34 & 1  & (1.3636, 0.9091) & 2  & (0.1270, 0.0928) & 3  & (3.8536, 3.0829) & 4  & (0.8722, 0.7631) & 5  & (0.2028, 0.1940) \\
& 6  & (0.5683, 0.5941) & 7  & (0.3832, 0.4379) & 8  & (0.2511, 0.3139) & 9  & (1.5501, 2.1213) & 10  & (9.2000, 13.8000) \\
    \midrule
    35 & 1  & (1.3636, 0.9091) & 2  & (13.5171, 9.8779) & 3  & (2.0712, 1.6569) & 4  & (3.7333, 3.2667) & 5  & (0.2983, 0.2854) \\
& 6  & (0.8046, 0.8412) & 7  & (0.1826, 0.2087) & 8  & (0.5052, 0.6315) & 9  & (0.3314, 0.4536) & 10  & (0.0800, 0.1200) \\
    \midrule
    36 & 1  & (0.2050, 0.1366) & 2  & (2.1213, 1.5501) & 3  & (3.8536, 3.0829) & 4  & (0.8722, 0.7631) & 5  & (0.4234, 0.4050) \\
& 6  & (1.1770, 1.2305) & 7  & (0.2695, 0.3080) & 8  & (10.5295, 13.1619) & 9  & (0.4691, 0.6420) & 10  & (0.0800, 0.1200) \\
    \midrule
    37 & 1  & (0.9247, 0.6165) & 2  & (0.1270, 0.0928) & 3  & (13.1619, 10.5295) & 4  & (0.4379, 0.3832) & 5  & (0.5941, 0.5683) \\
& 6  & (1.8504, 1.9345) & 7  & (0.1826, 0.2087) & 8  & (0.2511, 0.3139) & 9  & (2.8885, 3.9527) & 10  & (0.9091, 1.3636) \\
    \midrule
    38 & 1  & (0.6462, 0.4308) & 2  & (1.3443, 0.9823) & 3  & (0.8969, 0.7175) & 4  & (0.1304, 0.1141) & 5  & (0.2983, 0.2854) \\
& 6  & (3.4377, 3.5940) & 7  & (0.3832, 0.4379) & 8  & (10.5295, 13.1619) & 9  & (0.1536, 0.2101) & 10  & (1.4383, 2.1574) \\
    \midrule
    39 & 1  & (0.2050, 0.1366) & 2  & (0.6420, 0.4691) & 3  & (13.1619, 10.5295) & 4  & (2.0085, 1.7574) & 5  & (0.8412, 0.8046) \\
& 6  & (0.4050, 0.4234) & 7  & (0.2695, 0.3080) & 8  & (0.1043, 0.1303) & 9  & (2.8885, 3.9527) & 10  & (0.9091, 1.3636) \\
    \midrule
    40 & 1  & (13.8000, 9.2000) & 2  & (0.6420, 0.4691) & 3  & (0.1303, 0.1043) & 4  & (3.7333, 3.2667) & 5  & (0.4234, 0.4050) \\
& 6  & (1.8504, 1.9345) & 7  & (0.2695, 0.3080) & 8  & (0.7175, 0.8969) & 9  & (0.1536, 0.2101) & 10  & (0.9091, 1.3636) \\
    \midrule
    41 & 1  & (0.4537, 0.3024) & 2  & (3.9527, 2.8885) & 3  & (0.2113, 0.1690) & 4  & (0.6154, 0.5385) & 5  & (0.1275, 0.1219) \\
& 6  & (1.1770, 1.2305) & 7  & (11.1481, 12.7407) & 8  & (1.6569, 2.0712) & 9  & (0.6684, 0.9146) & 10  & (0.2085, 0.3127) \\
    \midrule
    42 & 1  & (1.3636, 0.9091) & 2  & (13.5171, 9.8779) & 3  & (2.0712, 1.6569) & 4  & (0.8722, 0.7631) & 5  & (3.5940, 3.4377) \\
& 6  & (0.2854, 0.2983) & 7  & (0.1826, 0.2087) & 8  & (0.1043, 0.1303) & 9  & (0.4691, 0.6420) & 10  & (0.3024, 0.4537) \\
    \midrule
    43 & 1  & (0.6462, 0.4308) & 2  & (13.5171, 9.8779) & 3  & (0.3139, 0.2511) & 4  & (1.2768, 1.1172) & 5  & (0.1275, 0.1219) \\
& 6  & (1.8504, 1.9345) & 7  & (0.3832, 0.4379) & 8  & (3.0829, 3.8536) & 9  & (0.6684, 0.9146) & 10  & (0.1366, 0.2050) \\
    \midrule
    44 & 1  & (1.3636, 0.9091) & 2  & (0.2101, 0.1536) & 3  & (0.4482, 0.3585) & 4  & (0.8722, 0.7631) & 5  & (0.2983, 0.2854) \\
& 6  & (1.8504, 1.9345) & 7  & (11.1481, 12.7407) & 8  & (3.0829, 3.8536) & 9  & (0.4691, 0.6420) & 10  & (0.0800, 0.1200) \\
    \midrule
    45 & 1  & (0.4537, 0.3024) & 2  & (3.9527, 2.8885) & 3  & (0.1303, 0.1043) & 4  & (1.2768, 1.1172) & 5  & (12.2604, 11.7273) \\
& 6  & (0.5683, 0.5941) & 7  & (1.7574, 2.0085) & 8  & (0.1690, 0.2113) & 9  & (0.6684, 0.9146) & 10  & (0.2085, 0.3127) \\
    \midrule
    46 & 1  & (0.1200, 0.0800) & 2  & (13.5171, 9.8779) & 3  & (3.8536, 3.0829) & 4  & (0.8722, 0.7631) & 5  & (0.2983, 0.2854) \\
& 6  & (1.8504, 1.9345) & 7  & (0.5385, 0.6154) & 8  & (0.1690, 0.2113) & 9  & (0.3314, 0.4536) & 10  & (0.9091, 1.3636) \\
    \midrule
    47 & 1  & (0.2050, 0.1366) & 2  & (0.1270, 0.0928) & 3  & (13.1619, 10.5295) & 4  & (1.2768, 1.1172) & 5  & (0.5941, 0.5683) \\
& 6  & (0.8046, 0.8412) & 7  & (3.2667, 3.7333) & 8  & (0.2511, 0.3139) & 9  & (0.3314, 0.4536) & 10  & (1.4383, 2.1574) \\
    \midrule
    48 & 1  & (13.8000, 9.2000) & 2  & (0.1270, 0.0928) & 3  & (0.6315, 0.5052) & 4  & (0.8722, 0.7631) & 5  & (1.2305, 1.1770) \\
& 6  & (0.4050, 0.4234) & 7  & (0.1826, 0.2087) & 8  & (1.6569, 2.0712) & 9  & (2.8885, 3.9527) & 10  & (0.2085, 0.3127) \\
    \midrule
    49 & 1  & (2.1574, 1.4383) & 2  & (0.4536, 0.3314) & 3  & (0.6315, 0.5052) & 4  & (0.2087, 0.1826) & 5  & (0.2983, 0.2854) \\
& 6  & (3.4377, 3.5940) & 7  & (1.1172, 1.2768) & 8  & (0.1043, 0.1303) & 9  & (0.6684, 0.9146) & 10  & (9.2000, 13.8000) \\
    \midrule
    50 & 1  & (13.8000, 9.2000) & 2  & (2.1213, 1.5501) & 3  & (0.2113, 0.1690) & 4  & (0.4379, 0.3832) & 5  & (0.1275, 0.1219) \\
& 6  & (0.5683, 0.5941) & 7  & (0.2695, 0.3080) & 8  & (1.0520, 1.3150) & 9  & (2.8885, 3.9527) & 10  & (0.6165, 0.9247) \\
\end{longtable}
\end{landscape}  

\section{Comparison Methods}\label{comparison_methods}

To evaluate the effectiveness of the proposed method, we selected several representative comparison methods. In this chapter, we provide an overview of these methods.

\begin{itemize}
    \item \textbf{Round-Robin:} A simple uniform sampling strategy that cycles through all arms regardless of observed outcomes. It serves as a fundamental baseline for fair but non-adaptive allocation, and is typically sample-inefficient.
    
    \item \textbf{Dominated Elimination Round-Robin (DE Round-Robin):} An enhanced version of Round-Robin that progressively eliminates arms empirically dominated in both mean and risk. It aims to reduce unnecessary sampling of clearly suboptimal arms.
    
    \item \textbf{Least-Important Elimination Round-Robin (LIE Round-Robin):} A fixed-budget strategy that eliminates arms contributing least to the current Pareto frontier. Initially allocates samples uniformly, then gradually focuses on promising arms.
    
    \item \textbf{Risk-Averse LUCB (RA-LUCB):} An extension of the classical LUCB algorithm to the risk-aware, multi-objective setting. It selects two arms each round—a potentially optimal arm and a challenger—and pulls both to reduce uncertainty near the Pareto frontier.
    
    \item \textbf{$\xi$ Lower Confidence Bound ($\xi$-LCB):} A Lower Confidence Bound (LCB)-based approach that targets arms with the lowest risk-adjusted performance, computed via the mean-variance trade-off. It encourages conservative exploration to identify risk-averse Pareto-optimal arms.
    
    \item \textbf{Hypervolume Improvement-based Pareto set Exploration (HVI-Pareto):} A method based on hypervolume improvement, selecting arms that most contribute to expanding the estimated Pareto front. This promotes diverse and well-distributed sampling across objectives. We set the reference point for hypervolume computation to $\left(R_\mu,R_\xi\right)=\left(0, \frac{0.25}{3+\rho}\right)$, which corresponds to the worst-case scenario under our problem setting: since rewards are scaled to $[0,1]$, the minimum possible mean is $0$ and the maximum possible variance is $0.25$. Thus, the maximum value of the risk measure $\xi = \alpha(\sigma^2 - \rho\mu)$ is $\frac{0.25}{3+\rho}$ when $\mu = 0$ and $\sigma^2 = 0.25$.
    
    \item \textbf{Empirical Gap-based Pareto Set Exploration (EGP):} EGP is a fixed-budget algorithm that aims to efficiently identify the Pareto-optimal arms by leveraging empirical dominance gaps. At each round, for each arm $i$, an empirical gap value $\widehat{V}_i(t)$ is computed as
\begin{equation*}
\widehat{V}_i(t) = 
\begin{cases}
\displaystyle \max_{j \neq i} \min\left( \widehat{\mu}_i(t) - \widehat{\mu}_j(t),\ \widehat{\xi}_j(t) - \widehat{\xi}_i(t) \right) & \text{if } i \in \widehat{D}_t^+, \\
\displaystyle \min_{j \neq i} \max\left( \widehat{\mu}_j(t) - \widehat{\mu}_i(t),\ \widehat{\xi}_i(t) - \widehat{\xi}_j(t) \right) & \text{if } i \notin \widehat{D}_t^+.
\end{cases}
\end{equation*}
The arm to be pulled is selected as $I(t) := \arg\max_i \left\{ -\widehat{V}_i(t) + \beta_i(t) \right\}$, where $\beta_i(t)$ denotes the confidence width for arm $i$. This sampling strategy focuses on arms close to the empirical Pareto frontier, effectively reducing uncertainty in the decision boundary and improving the quality of the final selection.
\end{itemize}

\begin{algorithm}[tb]
\rule{\linewidth}{1pt}
\caption{Round-Robin}
\label{alg:round-robin}
\vspace{-1.6ex}
\rule{\linewidth}{0.5pt}

\KwIn{$K$, $a$, $n$, $\epsilon$, $\rho$}

Pull each arm $i \in [K]$ twice and update $\hat{\mu}_i(t)$, $\hat{\xi}_i(t)$, $\underline{\mu}_i(t)$, $\overline{\mu}_i(t)$, $\underline{\xi}_i(t)$, $\overline{\xi}_i(t)$\;

Set $T_i(K) = 2$ for all $i$, and $t \gets 2K + 1$\;

\While{$t \le n$}{
    Identify $i_t \coloneq t \bmod K$\;

    Draw $X_{i_t}(T_{i_t}(t)+1) \sim \nu_{i_t}$\;

    $t \gets t + 1$\;

    Update $\hat{\mu}_{i_t}(t)$, $\hat{\xi}_{i_t}(t)$, $\beta_{i_t}(t)$, and $T_{i_t}(t)$\;

    \tcp*[l]{(Fixed-Budget Setting)}
    \If{$t > n$}{\textbf{break}}

    \tcp*[l]{(Fixed-Confidence Setting)}
    \If{$t > 2K$ \textbf{and} $V(t) < \epsilon$}{\textbf{break}}
}
\Return{$\widehat{D}^+_n$}

\vspace{-1.5ex}
\rule{\linewidth}{1pt}
\end{algorithm}

\begin{algorithm}[tb]
\rule{\linewidth}{1pt}
\caption{Dominated Elimination Round-Robin (DE Round-Robin)}
\label{alg:fc-rre}
\vspace{-1.6ex}
\rule{\linewidth}{0.5pt}

\KwIn{$K$, $\epsilon$, $\rho$}

Pull each arm $i \in [K]$ twice and update $\hat{\mu}_i(t)$, $\hat{\xi}_i(t)$, $\underline{\mu}_i(t)$, $\overline{\mu}_i(t)$, $\underline{\xi}_i(t)$, $\overline{\xi}_i(t)$\;

Set $T_i(K) = 2$ for all $i$, $t \gets 2K + 1$ and $I \gets 1$\;

\While{$V(t) > \epsilon$}{

    \For{$i = I \bmod K$}{

        \If{$i \in \widehat{D}^+_t\cup\left\{k\notin\widehat{D}^+_t|V_k(t)>\epsilon\right\}$}{
            Draw $X_i(T_i(t)+1) \sim \nu_i$\;
            
            Update $\hat{\mu}_i(t)$, $\hat{\xi}_i(t)$, $\underline{\mu}_i(t)$, $\overline{\mu}_i(t)$, $\underline{\xi}_i(t)$, $\overline{\xi}_i(t)$\;
            
            $t \gets t + 1$\;
            
            $I \gets I + 1$\;
        }

        \Else{
            $I \gets I + 1$\;
        }
    }
}

\Return{$\widehat{D}^+_{\widetilde{n}}$}

\vspace{-1.5ex}
\rule{\linewidth}{1pt}
\end{algorithm}

\begin{algorithm}[tb]
\rule{\linewidth}{1pt}
\caption{Least-important elimination Round-Robin (LIE Round-Robin)}
\label{alg:lie-rr}
\vspace{-1.6ex}
\rule{\linewidth}{0.5pt}

\KwIn{$K$, $a$, $n$, $\epsilon$, $\rho$}

Pull each arm $i \in [K]$ twice and update $\hat{\mu}_i(t)$, $\hat{\xi}_i(t)$, $\underline{\mu}_i(t)$, $\overline{\mu}_i(t)$, $\underline{\xi}_i(t)$, $\overline{\xi}_i(t)$\;

Set $T_i(K) = 2$ for all $i$, $t \gets 2K + 1$ and $I \gets 1$\;

\While{$t \le n$}{
    Identify $i_I \coloneq I \bmod K$\;

    \If{$i_I \neq \arg\min_{i \in [K]} V_i(t)$}{
        Draw $X_{i_I}(T_{i_I}(t)+1) \sim \nu_{i_I}$\;

        $t \gets t + 1$\; 
        
        $I \gets I + 1$\;

        Update $\hat{\mu}_{i_I}(t)$, $\hat{\xi}_{i_I}(t)$, $\beta_{i_I}(t)$, and $T_{i_I}(t)$\;
    }
    \Else{
        $I \gets I + 1$\;
    }

    \If{$t > n$}{\textbf{break}}

}

\Return{$\widehat{D}^+_n$}

\vspace{-1.5ex}
\rule{\linewidth}{1pt}
\end{algorithm}

\begin{algorithm}[tb]
\rule{\linewidth}{1pt}
\caption{Risk-Averse Lower and Upper Confidence Bounds (RA-LUCB)}
\label{alg:ra-lucb}
\vspace{-1.6ex}
\rule{\linewidth}{0.5pt}

\KwIn{$K$, $a$, $n$, $\epsilon$, $\rho$}

Pull each arm $i \in [K]$ twice and update $\hat{\mu}_i(t)$, $\hat{\xi}_i(t)$, $\underline{\mu}_i(t)$, $\overline{\mu}_i(t)$, $\underline{\xi}_i(t)$, $\overline{\xi}_i(t)$\;

Set $T_i(K) = 2$ for all $i$ and $t \gets 2K + 1$\;

\While{$t \le n$}{

    Identify $m_t$ and $p_t$ by Eq.~\ref{mt} and Eq.~\ref{pt}\;

    Draw $X_{m_t}(T_{m_t}(t)+1) \sim \nu_{m_t}$\;
    
    Draw $X_{p_t}(T_{p_t}(t)+1) \sim \nu_{p_t}$\;

    $t \gets t + 2$\;

    Update $\hat{\mu}_{m_t}(t)$, $\hat{\mu}_{p_t}(t)$, $\hat{\xi}_{m_t}(t)$, $\hat{\xi}_{p_t}(t)$, $\beta_{m_t}(t)$, $\beta_{p_t}(t)$, $T_{m_t}(t)$, and $T_{p_t}(t)$\;

    \tcp*[l]{(Fixed-Budget Setting)}
    \If{$t > n$}{\textbf{break}}

    \tcp*[l]{(Fixed-Confidence Setting)}
    \If{$t > 2K \wedge V(t) < \epsilon$}{\textbf{break}}
}
\Return{$\widehat{D}^+_n$}

\vspace{-1.5ex}
\rule{\linewidth}{1pt}
\end{algorithm}

\begin{algorithm}[tb]
\rule{\linewidth}{1pt}
\caption{$\xi$ Lower Confidence Bound ($\xi$-LCB)}
\label{alg:xi-lcb}
\vspace{-1.6ex}
\rule{\linewidth}{0.5pt}

\KwIn{$K$, $a$, $n$, $\epsilon$, $\rho$}

Pull each arm $i \in [K]$ twice and update $\hat{\mu}_i(t)$, $\hat{\xi}_i(t)$, $\underline{\mu}_i(t)$, $\overline{\mu}_i(t)$, $\underline{\xi}_i(t)$, $\overline{\xi}_i(t)$\;

Set $T_i(K) = 2$ for all $i$ and $t \gets 2K + 1$\;

\While{$t \le n$}{

    Identify $i_t \coloneq \arg\min_{i \in [K]} \underline{\xi}_i(t)$\;

    Draw $X_{i_t}(T_{i_t}(t)+1) \sim \nu_{i_t}$\;

    $t \gets t + 1$\;

    Update $\hat{\mu}_{i_t}(t)$, $\hat{\xi}_{i_t}(t)$, $\beta_{i_t}(t)$, and $T_{i_t}(t)$\;

    \If{$t > n$}{\textbf{break}}

}
\Return{$\widehat{D}^+_n$}

\vspace{-1.5ex}
\rule{\linewidth}{1pt}
\end{algorithm}

\begin{algorithm}[tb]
\rule{\linewidth}{1pt}
\caption{Hypervolume Improvement-based Pareto set Exploration (HVI-Pareto)}
\label{alg:hv-pareto}
\vspace{-1.6ex}
\rule{\linewidth}{0.5pt}

\KwIn{$K$, $a$, $n$, $\epsilon$, $\rho$, reference point $(R_\mu, R_\xi)$}

Pull each arm $i \in [K]$ twice and update $\hat{\mu}_i(t)$, $\hat{\xi}_i(t)$, $\underline{\mu}_i(t)$, $\overline{\mu}_i(t)$, $\underline{\xi}_i(t)$, $\overline{\xi}_i(t)$\;

Set $T_i(K) = 2$ for all $i$ and $t \gets 2K + 1$\;

\While{$t \le n$}{
    Compute $HVI_i(t) \coloneq (\hat{\mu}_i(t) - R_\mu)(R_\xi - \hat{\xi}_i(t))$ for each $i$\;

    Identify $d_t \coloneq \arg\min_{i \in \widehat{D}^+_t} HVI_i(t)$ and $d_t' \coloneq \arg\max_{i \notin \widehat{D}^+_t} HVI_i(t)$\;

    Draw $X_{d_t}(T_{d_t}(t)+1) \sim \nu_{d_t}$\;

    Draw $X_{d_t'}(T_{d_t'}(t)+1) \sim \nu_{d_t'}$\;

    $t \gets t + 2$\;

    Update $\hat{\mu}_{d_t}(t)$, $\hat{\mu}_{d_t'}(t)$, $\hat{\xi}_{d_t}(t)$, $\hat{\xi}_{d_t'}(t)$, $\beta_{d_t}(t)$, $\beta_{d_t'}(t)$, $T_{d_t}(t)$, and $T_{d_t'}(t)$\;

    \If{$t > n$}{\textbf{break}}

}
\Return{$\widehat{D}^+_n$}

\vspace{-1.5ex}
\rule{\linewidth}{1pt}
\end{algorithm}

\begin{algorithm}[tb]
\rule{\linewidth}{1pt}
\caption{Empirical Gap-based Pareto Set Exploration (EGP)}
\label{alg:egp}
\vspace{-1.6ex}
\rule{\linewidth}{0.5pt}

\KwIn{$K$, $a$, $n$, $\epsilon$, $\rho$}

Pull each arm $i \in [K]$ twice and update $\hat{\mu}_i(t)$, $\hat{\xi}_i(t)$, $\underline{\mu}_i(t)$, $\overline{\mu}_i(t)$, $\underline{\xi}_i(t)$, $\overline{\xi}_i(t)$\;

Set $T_i(K) = 2$ for all $i$ and $t \gets 2K + 1$\;

\While{$t \le n$}{
    Compute $ \widehat{V}_i(t) = \begin{cases} 
  \underset{j\neq i}{\max}\ \min\left(\hat{\mu}_i(t)-\hat{\mu}_j(t),\hat{\xi}_j(t)-\hat{\xi}_i(t)\right)& \text{if }i\in\widehat{D}^+_t \\
  \underset{j\neq i}{\min}\ \max\left(\hat{\mu}_j(t)-\hat{\mu}_i(t),\hat{\xi}_i(t)-\hat{\xi}_j(t)\right) & \text{if }i\notin\widehat{D}^+_t \end{cases},$ for $i\in[K]$

  Identify $I(t)\coloneq\underset{i\in[K]}{\operatorname{argmax}}\ \left(-\widehat{V}_i(t)+\beta_i(t)\right)$

  Draw $X_{I(t)}(T_{I(t)}(t)+1) \sim \nu_{I(t)}$\;

    $t \gets t + 1$\;

    Update $\hat{\mu}_{I(t)}(t)$, $\hat{\xi}_{I(t)}(t)$, $\beta_{I(t)}(t)$, and $T_{I(t)}(t)$\;



    \If{$t > n$}{\textbf{break}}

}

\Return{$\widehat{D}^+_n$}

\vspace{-1.5ex}
\rule{\linewidth}{1pt}
\end{algorithm}
\clearpage
\section{Additional Experimental Results}\label{sec:appendix_additional_results}
\subsection{Performance under Unbounded Gaussian Rewards}\label{sec:gaussian}
In this section, we provide additional experimental results to further demonstrate performance and robustness when rewards are not bounded in $[0,1]$.  Here, we replicate Experiment 1 (Stopping Time Comparison) using Gaussian rewards.
\vskip\baselineskip
\noindent
For each arm $i\in[K]$, the reward distribution is given by 
\begin{equation*}
    X_i\sim\mathcal{N}(\mu_i, \sigma_i^2),
\end{equation*}
where the mean $\mu_i$ is in $[0.4,0.6]$ and the variance $\sigma^2_i$ is in $[0.01,0.2]$, consistent with the setting of the original Beta distribution experiments.

\subsubsection*{Confidence Interval for Mean, Variance, and MV}

For unbounded Gaussian rewards, we derive high-probability confidence intervals for 
mean and variance. In the following, we employ a Hoeffding-type union bound over time that is less tight than what we presented for bounded rewards in the main text (Note that
tighter bounds can further improve the RAMGapE performance). 
Here, we applied the concentration inequalities introduced in~\cite{wainwright2019high}, in which the 
mean is treated using the standard sub-Gaussian concentration inequality and for the 
variance we used the centered squared deviations $Y_i\coloneq(X_i-\mu_i)^2-\sigma_i^2$, and $(\nu^2,B)$-sub-exponential property.
Following~\cite{honorio2014tight}, we adopt the sub-exponential parameters $(\nu,B)=(4\sqrt{2}\sigma_i^2,4\sigma_i^2)$ when applying the Bernstein-type inequality to derive the confidence interval for the variance.

As results, for each arm $i$, with probability at least $1-\delta$, the following confidence intervals are obtained for mean $\mu_i$ and variance $\sigma_i^2$:

\begin{eqnarray*}
    |\hat{\mu}_i(t)-\mu_i|
    &\leq&
    \sigma_\text{max}\sqrt{\frac{2}{T_i(t)}\ln\frac{4KT_i^2(t)}{\delta}},\\
    |\hat{\sigma}^2_i(t)-\sigma^2_i|
    &\leq&
8\sigma_\text{max}^2\max\left(\sqrt{\frac{1}{T_i(t)}\ln\frac{4KT_i^2(t)}{\delta}},\frac{1}{T_i(t)}\ln\frac{4KT_i^2(t)}{\delta}\right)+\frac{2\sigma_\text{max}^2}{T_i(t)}\ln\frac{4KT_i^2(t)}{\delta},
\end{eqnarray*}
where $\hat{\mu}_i(t)$ and $\hat{\sigma}_i^2(t)$ are the empirical mean and variance over $T_i(t)$ samples. In the derivation, the formal solutions are acquired in terms of the true $\sigma_i$ at the location of $\sigma_{\text{max}}$ in the right-hand sides. Because the true $\sigma_i$ is unknown a priori to the algorithm, we replace it with a possible maximum value of $\sigma_i$, keeping the property of upper bounds. In the current experimental setup, we can simply use $\sigma_{\text{max}}=\sqrt{0.2}$.

Finally, we can derive the corresponding confidence interval for $\text{MV}$ for Gaussian rewards. For each arm $i$ and for each round $t$, we denote the upper and lower bounds of $\text{MV}_i$ as
$\overline{\text{MV}}_i(t)\coloneq\overline{\sigma}_i(t)-\rho\underline{\mu}_i(t)$ and $\underline{\text{MV}}_i(t)\coloneq\underline{\sigma}^2_i(t)-\rho\overline{\mu}_i(t)$ (recall $\rho>0$), where the underlines and overlines denote the lower and upper confidence bounds of the respective quantities.

\subsubsection*{Experiment Part}

We replicate Experiment 1 (Stopping Time Comparison) under Gaussian rewards. The setup is identical to that of Experiment 1 with Beta distributions, except for the reward generation process. Specifically, we conduct simulations across 50 problem instances, each comprising $K=10$ arms. For each instance, the mean and variance of arms are drawn from the patterns detailed in Table~\ref{table_gaussian_patterns}. The algorithmic parameters are fixed at $(\delta,\epsilon,\rho)=(0.05,0.1,0.01)$.

\begin{figure}[t]
  \centering

  \subfigure[vs RA-LUCB]{%
    \includegraphics[width=0.325\textwidth]{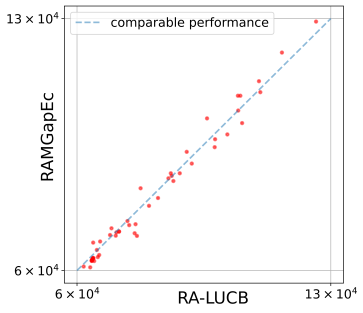}}
  \hfill
  \subfigure[vs DE Round-Robin]{%
    \includegraphics[width=0.325\textwidth]{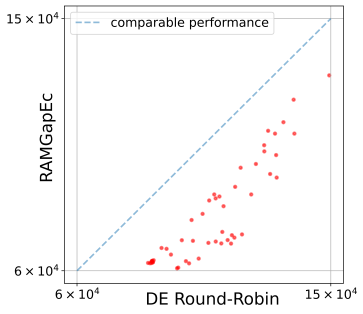}}
    \hfill
    \subfigure[vs Round-Robin]{%
    \includegraphics[width=0.325\textwidth]{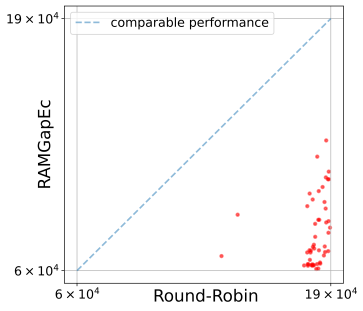}}
  \caption{Stopping Time Comparison of Experiment 1 with  $(\delta,\epsilon,\rho)=(0.05,0.1,0.01)$ for Gaussian rewards. The blue dashed line corresponds to the identity line, i.e., the set of points where both methods terminate at the same time, indicating comparable performance. Points located below this line signify that the proposed method stops earlier than the baseline.}
  \label{fig:gaussian_stopping_time}
\end{figure}

The results are shown in Fig.~\ref{fig:gaussian_stopping_time}. The plot demonstrates a trend consistent with our findings for Beta rewards: RAMGapEc terminates significantly faster than the Round-Robin-based baseline algorithms, and as fast as RA-LUCB. This result supports that RAMGapE is a robust algorithm whose effectiveness is not necessarily confined to bounded reward settings.

\begin{landscape}
\begin{longtable}{c||c c c c c c c c c c c}
    \caption{Mean ($\mu$) and Variance ($\sigma^2$) values of each pattern for Gaussian distribution experiments} \label{table_gaussian_patterns}\\
    \toprule
    pattern & \multicolumn{10}{c}{Index $(\mu, \sigma^2)$}  \\
    \midrule
    \endfirsthead

    \toprule
    pattern & \multicolumn{10}{c}{Index $(\mu, \sigma^2)$} \\
    \midrule
    \endhead

    \midrule
    \multicolumn{11}{r}{Go to next page} \\
    \midrule
    \endfoot

    \bottomrule
    \endlastfoot
    \midrule
    1  & 1  & (0.6000, 0.0522) & 2  & (0.5778, 0.1789) & 3  & (0.5556, 0.0944) & 4  & (0.5333, 0.2000) & 5  & (0.5111, 0.0311) \\
& 6  & (0.4889, 0.1367) & 7  & (0.4667, 0.1578) & 8  & (0.4444, 0.0733) & 9  & (0.4222, 0.0100) & 10  & (0.4000, 0.1156) \\
    \midrule
    2  & 1  & (0.6000, 0.0944) & 2  & (0.5778, 0.1156) & 3  & (0.5556, 0.1578) & 4  & (0.5333, 0.0311) & 5  & (0.5111, 0.0522) \\
& 6  & (0.4889, 0.2000) & 7  & (0.4667, 0.1789) & 8  & (0.4444, 0.0733) & 9  & (0.4222, 0.0100) & 10  & (0.4000, 0.1367) \\
    \midrule
    3  & 1  & (0.6000, 0.0522) & 2  & (0.5778, 0.1156) & 3  & (0.5556, 0.2000) & 4  & (0.5333, 0.0944) & 5  & (0.5111, 0.0733) \\
& 6  & (0.4889, 0.1578) & 7  & (0.4667, 0.0311) & 8  & (0.4444, 0.1789) & 9  & (0.4222, 0.1367) & 10  & (0.4000, 0.0100) \\
    \midrule
    4  & 1  & (0.6000, 0.1578) & 2  & (0.5778, 0.0733) & 3  & (0.5556, 0.1156) & 4  & (0.5333, 0.2000) & 5  & (0.5111, 0.0100) \\
& 6  & (0.4889, 0.0311) & 7  & (0.4667, 0.1789) & 8  & (0.4444, 0.0522) & 9  & (0.4222, 0.0944) & 10  & (0.4000, 0.1367) \\
    \midrule
    5  & 1  & (0.6000, 0.2000) & 2  & (0.5778, 0.0944) & 3  & (0.5556, 0.0100) & 4  & (0.5333, 0.1367) & 5  & (0.5111, 0.1156) \\
& 6  & (0.4889, 0.1789) & 7  & (0.4667, 0.1578) & 8  & (0.4444, 0.0733) & 9  & (0.4222, 0.0311) & 10  & (0.4000, 0.0522) \\
    \midrule
    6  & 1  & (0.6000, 0.0522) & 2  & (0.5778, 0.1789) & 3  & (0.5556, 0.0100) & 4  & (0.5333, 0.1156) & 5  & (0.5111, 0.0733) \\
& 6  & (0.4889, 0.0944) & 7  & (0.4667, 0.2000) & 8  & (0.4444, 0.0311) & 9  & (0.4222, 0.1578) & 10  & (0.4000, 0.1367) \\
    \midrule
    7  & 1  & (0.6000, 0.1578) & 2  & (0.5778, 0.1789) & 3  & (0.5556, 0.0311) & 4  & (0.5333, 0.0522) & 5  & (0.5111, 0.2000) \\
& 6  & (0.4889, 0.0944) & 7  & (0.4667, 0.0100) & 8  & (0.4444, 0.0733) & 9  & (0.4222, 0.1156) & 10  & (0.4000, 0.1367) \\
    \midrule
    8  & 1  & (0.6000, 0.1156) & 2  & (0.5778, 0.0944) & 3  & (0.5556, 0.1578) & 4  & (0.5333, 0.0311) & 5  & (0.5111, 0.1789) \\
& 6  & (0.4889, 0.1367) & 7  & (0.4667, 0.0733) & 8  & (0.4444, 0.0522) & 9  & (0.4222, 0.0100) & 10  & (0.4000, 0.2000) \\
    \midrule
    9  & 1  & (0.6000, 0.0100) & 2  & (0.5778, 0.0733) & 3  & (0.5556, 0.2000) & 4  & (0.5333, 0.1156) & 5  & (0.5111, 0.1578) \\
& 6  & (0.4889, 0.1367) & 7  & (0.4667, 0.0522) & 8  & (0.4444, 0.0944) & 9  & (0.4222, 0.1789) & 10  & (0.4000, 0.0311) \\
    \midrule
    10 & 1  & (0.6000, 0.1789) & 2  & (0.5778, 0.1578) & 3  & (0.5556, 0.0944) & 4  & (0.5333, 0.2000) & 5  & (0.5111, 0.0733) \\
& 6  & (0.4889, 0.0311) & 7  & (0.4667, 0.1156) & 8  & (0.4444, 0.0100) & 9  & (0.4222, 0.0522) & 10  & (0.4000, 0.1367) \\
    \midrule
    11 & 1  & (0.6000, 0.0522) & 2  & (0.5778, 0.0944) & 3  & (0.5556, 0.0311) & 4  & (0.5333, 0.1156) & 5  & (0.5111, 0.2000) \\
& 6  & (0.4889, 0.1578) & 7  & (0.4667, 0.1789) & 8  & (0.4444, 0.0100) & 9  & (0.4222, 0.0733) & 10  & (0.4000, 0.1367) \\
    \midrule
    12 & 1  & (0.6000, 0.0100) & 2  & (0.5778, 0.0733) & 3  & (0.5556, 0.0311) & 4  & (0.5333, 0.1789) & 5  & (0.5111, 0.2000) \\
& 6  & (0.4889, 0.1578) & 7  & (0.4667, 0.0944) & 8  & (0.4444, 0.1156) & 9  & (0.4222, 0.0522) & 10  & (0.4000, 0.1367) \\
    \midrule
    13 & 1  & (0.6000, 0.1578) & 2  & (0.5778, 0.0522) & 3  & (0.5556, 0.1156) & 4  & (0.5333, 0.0100) & 5  & (0.5111, 0.2000) \\
& 6  & (0.4889, 0.1367) & 7  & (0.4667, 0.1789) & 8  & (0.4444, 0.0311) & 9  & (0.4222, 0.0733) & 10  & (0.4000, 0.0944) \\
    \midrule
    14 & 1  & (0.6000, 0.0100) & 2  & (0.5778, 0.1367) & 3  & (0.5556, 0.1789) & 4  & (0.5333, 0.0522) & 5  & (0.5111, 0.2000) \\
& 6  & (0.4889, 0.0311) & 7  & (0.4667, 0.0733) & 8  & (0.4444, 0.0944) & 9  & (0.4222, 0.1578) & 10  & (0.4000, 0.1156) \\
    \midrule
    15 & 1  & (0.6000, 0.0522) & 2  & (0.5778, 0.1156) & 3  & (0.5556, 0.0100) & 4  & (0.5333, 0.0311) & 5  & (0.5111, 0.2000) \\
& 6  & (0.4889, 0.1789) & 7  & (0.4667, 0.1367) & 8  & (0.4444, 0.0944) & 9  & (0.4222, 0.0733) & 10  & (0.4000, 0.1578) \\
    \midrule
    16 & 1  & (0.6000, 0.0100) & 2  & (0.5778, 0.1367) & 3  & (0.5556, 0.1156) & 4  & (0.5333, 0.0311) & 5  & (0.5111, 0.0944) \\
& 6  & (0.4889, 0.0522) & 7  & (0.4667, 0.1578) & 8  & (0.4444, 0.2000) & 9  & (0.4222, 0.1789) & 10  & (0.4000, 0.0733) \\
    \midrule
    17 & 1  & (0.6000, 0.1578) & 2  & (0.5778, 0.1156) & 3  & (0.5556, 0.0100) & 4  & (0.5333, 0.2000) & 5  & (0.5111, 0.1789) \\
& 6  & (0.4889, 0.0944) & 7  & (0.4667, 0.0311) & 8  & (0.4444, 0.1367) & 9  & (0.4222, 0.0522) & 10  & (0.4000, 0.0733) \\
    \midrule
    18 & 1  & (0.6000, 0.1367) & 2  & (0.5778, 0.0100) & 3  & (0.5556, 0.0944) & 4  & (0.5333, 0.2000) & 5  & (0.5111, 0.1789) \\
& 6  & (0.4889, 0.1578) & 7  & (0.4667, 0.0733) & 8  & (0.4444, 0.0522) & 9  & (0.4222, 0.0311) & 10  & (0.4000, 0.1156) \\
    \midrule
    19 & 1  & (0.6000, 0.0522) & 2  & (0.5778, 0.1156) & 3  & (0.5556, 0.1367) & 4  & (0.5333, 0.1789) & 5  & (0.5111, 0.0944) \\
& 6  & (0.4889, 0.0100) & 7  & (0.4667, 0.0733) & 8  & (0.4444, 0.1578) & 9  & (0.4222, 0.0311) & 10  & (0.4000, 0.2000) \\
    \midrule
    20 & 1  & (0.6000, 0.1156) & 2  & (0.5778, 0.1578) & 3  & (0.5556, 0.2000) & 4  & (0.5333, 0.0733) & 5  & (0.5111, 0.0311) \\
& 6  & (0.4889, 0.0944) & 7  & (0.4667, 0.1789) & 8  & (0.4444, 0.0522) & 9  & (0.4222, 0.1367) & 10  & (0.4000, 0.0100) \\
    \midrule
    21 & 1  & (0.6000, 0.0522) & 2  & (0.5778, 0.1578) & 3  & (0.5556, 0.1367) & 4  & (0.5333, 0.0944) & 5  & (0.5111, 0.1156) \\
& 6  & (0.4889, 0.2000) & 7  & (0.4667, 0.1789) & 8  & (0.4444, 0.0100) & 9  & (0.4222, 0.0311) & 10  & (0.4000, 0.0733) \\
    \midrule
    22 & 1  & (0.6000, 0.1578) & 2  & (0.5778, 0.1367) & 3  & (0.5556, 0.0100) & 4  & (0.5333, 0.0944) & 5  & (0.5111, 0.2000) \\
& 6  & (0.4889, 0.1789) & 7  & (0.4667, 0.0522) & 8  & (0.4444, 0.1156) & 9  & (0.4222, 0.0311) & 10  & (0.4000, 0.0733) \\
    \midrule
    23 & 1  & (0.6000, 0.0100) & 2  & (0.5778, 0.1367) & 3  & (0.5556, 0.0733) & 4  & (0.5333, 0.0522) & 5  & (0.5111, 0.0944) \\
& 6  & (0.4889, 0.1156) & 7  & (0.4667, 0.1578) & 8  & (0.4444, 0.0311) & 9  & (0.4222, 0.2000) & 10  & (0.4000, 0.1789) \\
    \midrule
    24 & 1  & (0.6000, 0.1156) & 2  & (0.5778, 0.2000) & 3  & (0.5556, 0.0733) & 4  & (0.5333, 0.1789) & 5  & (0.5111, 0.0944) \\
& 6  & (0.4889, 0.0311) & 7  & (0.4667, 0.1367) & 8  & (0.4444, 0.0100) & 9  & (0.4222, 0.1578) & 10  & (0.4000, 0.0522) \\
    \midrule
    25 & 1  & (0.6000, 0.0522) & 2  & (0.5778, 0.0944) & 3  & (0.5556, 0.1578) & 4  & (0.5333, 0.0100) & 5  & (0.5111, 0.0311) \\
& 6  & (0.4889, 0.1367) & 7  & (0.4667, 0.2000) & 8  & (0.4444, 0.1156) & 9  & (0.4222, 0.1789) & 10  & (0.4000, 0.0733) \\
    \midrule
    26 & 1  & (0.6000, 0.1367) & 2  & (0.5778, 0.0944) & 3  & (0.5556, 0.0100) & 4  & (0.5333, 0.1789) & 5  & (0.5111, 0.0522) \\
& 6  & (0.4889, 0.1578) & 7  & (0.4667, 0.0733) & 8  & (0.4444, 0.1156) & 9  & (0.4222, 0.2000) & 10  & (0.4000, 0.0311) \\
    \midrule
    27 & 1  & (0.6000, 0.2000) & 2  & (0.5778, 0.0522) & 3  & (0.5556, 0.0311) & 4  & (0.5333, 0.1156) & 5  & (0.5111, 0.0100) \\
& 6  & (0.4889, 0.1789) & 7  & (0.4667, 0.0944) & 8  & (0.4444, 0.1367) & 9  & (0.4222, 0.0733) & 10  & (0.4000, 0.1578) \\
    \midrule
    28 & 1  & (0.6000, 0.0100) & 2  & (0.5778, 0.0311) & 3  & (0.5556, 0.1578) & 4  & (0.5333, 0.0522) & 5  & (0.5111, 0.1789) \\
& 6  & (0.4889, 0.0944) & 7  & (0.4667, 0.1367) & 8  & (0.4444, 0.2000) & 9  & (0.4222, 0.0733) & 10  & (0.4000, 0.1156) \\
    \midrule
    29 & 1  & (0.6000, 0.1156) & 2  & (0.5778, 0.1578) & 3  & (0.5556, 0.0733) & 4  & (0.5333, 0.2000) & 5  & (0.5111, 0.0522) \\
& 6  & (0.4889, 0.0100) & 7  & (0.4667, 0.1789) & 8  & (0.4444, 0.1367) & 9  & (0.4222, 0.0944) & 10  & (0.4000, 0.0311) \\
    \midrule
    30 & 1  & (0.6000, 0.0944) & 2  & (0.5778, 0.1789) & 3  & (0.5556, 0.1578) & 4  & (0.5333, 0.1367) & 5  & (0.5111, 0.1156) \\
& 6  & (0.4889, 0.0733) & 7  & (0.4667, 0.0311) & 8  & (0.4444, 0.0522) & 9  & (0.4222, 0.2000) & 10  & (0.4000, 0.0100) \\
    \midrule
    31 & 1  & (0.6000, 0.0944) & 2  & (0.5778, 0.1789) & 3  & (0.5556, 0.1578) & 4  & (0.5333, 0.1367) & 5  & (0.5111, 0.0311) \\
& 6  & (0.4889, 0.1156) & 7  & (0.4667, 0.0522) & 8  & (0.4444, 0.2000) & 9  & (0.4222, 0.0100) & 10  & (0.4000, 0.0733) \\
    \midrule
    32 & 1  & (0.6000, 0.0311) & 2  & (0.5778, 0.1789) & 3  & (0.5556, 0.1367) & 4  & (0.5333, 0.0100) & 5  & (0.5111, 0.1156) \\
& 6  & (0.4889, 0.0733) & 7  & (0.4667, 0.0522) & 8  & (0.4444, 0.0944) & 9  & (0.4222, 0.2000) & 10  & (0.4000, 0.1578) \\
    \midrule
    33 & 1  & (0.6000, 0.1578) & 2  & (0.5778, 0.1789) & 3  & (0.5556, 0.1367) & 4  & (0.5333, 0.2000) & 5  & (0.5111, 0.0100) \\
& 6  & (0.4889, 0.0311) & 7  & (0.4667, 0.1156) & 8  & (0.4444, 0.0522) & 9  & (0.4222, 0.0733) & 10  & (0.4000, 0.0944) \\
    \midrule
    34 & 1  & (0.6000, 0.0733) & 2  & (0.5778, 0.2000) & 3  & (0.5556, 0.0311) & 4  & (0.5333, 0.0944) & 5  & (0.5111, 0.1789) \\
& 6  & (0.4889, 0.1156) & 7  & (0.4667, 0.1367) & 8  & (0.4444, 0.1578) & 9  & (0.4222, 0.0522) & 10  & (0.4000, 0.0100) \\
    \midrule
    35 & 1  & (0.6000, 0.0733) & 2  & (0.5778, 0.0100) & 3  & (0.5556, 0.0522) & 4  & (0.5333, 0.0311) & 5  & (0.5111, 0.1578) \\
& 6  & (0.4889, 0.0944) & 7  & (0.4667, 0.1789) & 8  & (0.4444, 0.1156) & 9  & (0.4222, 0.1367) & 10  & (0.4000, 0.2000) \\
    \midrule
    36 & 1  & (0.6000, 0.1789) & 2  & (0.5778, 0.0522) & 3  & (0.5556, 0.0311) & 4  & (0.5333, 0.0944) & 5  & (0.5111, 0.1367) \\
& 6  & (0.4889, 0.0733) & 7  & (0.4667, 0.1578) & 8  & (0.4444, 0.0100) & 9  & (0.4222, 0.1156) & 10  & (0.4000, 0.2000) \\
    \midrule
    37 & 1  & (0.6000, 0.0944) & 2  & (0.5778, 0.2000) & 3  & (0.5556, 0.0100) & 4  & (0.5333, 0.1367) & 5  & (0.5111, 0.1156) \\
& 6  & (0.4889, 0.0522) & 7  & (0.4667, 0.1789) & 8  & (0.4444, 0.1578) & 9  & (0.4222, 0.0311) & 10  & (0.4000, 0.0733) \\
    \midrule
    38 & 1  & (0.6000, 0.1156) & 2  & (0.5778, 0.0733) & 3  & (0.5556, 0.0944) & 4  & (0.5333, 0.2000) & 5  & (0.5111, 0.1578) \\
& 6  & (0.4889, 0.0311) & 7  & (0.4667, 0.1367) & 8  & (0.4444, 0.0100) & 9  & (0.4222, 0.1789) & 10  & (0.4000, 0.0522) \\
    \midrule
    39 & 1  & (0.6000, 0.1789) & 2  & (0.5778, 0.1156) & 3  & (0.5556, 0.0100) & 4  & (0.5333, 0.0522) & 5  & (0.5111, 0.0944) \\
& 6  & (0.4889, 0.1367) & 7  & (0.4667, 0.1578) & 8  & (0.4444, 0.2000) & 9  & (0.4222, 0.0311) & 10  & (0.4000, 0.0733) \\
    \midrule
    40 & 1  & (0.6000, 0.0100) & 2  & (0.5778, 0.1156) & 3  & (0.5556, 0.2000) & 4  & (0.5333, 0.0311) & 5  & (0.5111, 0.1367) \\
& 6  & (0.4889, 0.0522) & 7  & (0.4667, 0.1578) & 8  & (0.4444, 0.0944) & 9  & (0.4222, 0.1789) & 10  & (0.4000, 0.0733) \\
    \midrule
    41 & 1  & (0.6000, 0.1367) & 2  & (0.5778, 0.0311) & 3  & (0.5556, 0.1789) & 4  & (0.5333, 0.1156) & 5  & (0.5111, 0.2000) \\
& 6  & (0.4889, 0.0733) & 7  & (0.4667, 0.0100) & 8  & (0.4444, 0.0522) & 9  & (0.4222, 0.0944) & 10  & (0.4000, 0.1578) \\
    \midrule
    42 & 1  & (0.6000, 0.0733) & 2  & (0.5778, 0.0100) & 3  & (0.5556, 0.0522) & 4  & (0.5333, 0.0944) & 5  & (0.5111, 0.0311) \\
& 6  & (0.4889, 0.1578) & 7  & (0.4667, 0.1789) & 8  & (0.4444, 0.2000) & 9  & (0.4222, 0.1156) & 10  & (0.4000, 0.1367) \\
    \midrule
    43 & 1  & (0.6000, 0.1156) & 2  & (0.5778, 0.0100) & 3  & (0.5556, 0.1578) & 4  & (0.5333, 0.0733) & 5  & (0.5111, 0.2000) \\
& 6  & (0.4889, 0.0522) & 7  & (0.4667, 0.1367) & 8  & (0.4444, 0.0311) & 9  & (0.4222, 0.0944) & 10  & (0.4000, 0.1789) \\
    \midrule
    44 & 1  & (0.6000, 0.0733) & 2  & (0.5778, 0.1789) & 3  & (0.5556, 0.1367) & 4  & (0.5333, 0.0944) & 5  & (0.5111, 0.1578) \\
& 6  & (0.4889, 0.0522) & 7  & (0.4667, 0.0100) & 8  & (0.4444, 0.0311) & 9  & (0.4222, 0.1156) & 10  & (0.4000, 0.2000) \\
    \midrule
    45 & 1  & (0.6000, 0.1367) & 2  & (0.5778, 0.0311) & 3  & (0.5556, 0.2000) & 4  & (0.5333, 0.0733) & 5  & (0.5111, 0.0100) \\
& 6  & (0.4889, 0.1156) & 7  & (0.4667, 0.0522) & 8  & (0.4444, 0.1789) & 9  & (0.4222, 0.0944) & 10  & (0.4000, 0.1578) \\
    \midrule
    46 & 1  & (0.6000, 0.2000) & 2  & (0.5778, 0.0100) & 3  & (0.5556, 0.0311) & 4  & (0.5333, 0.0944) & 5  & (0.5111, 0.1578) \\
& 6  & (0.4889, 0.0522) & 7  & (0.4667, 0.1156) & 8  & (0.4444, 0.1789) & 9  & (0.4222, 0.1367) & 10  & (0.4000, 0.0733) \\
    \midrule
    47 & 1  & (0.6000, 0.1789) & 2  & (0.5778, 0.2000) & 3  & (0.5556, 0.0100) & 4  & (0.5333, 0.0733) & 5  & (0.5111, 0.1156) \\
& 6  & (0.4889, 0.0944) & 7  & (0.4667, 0.0311) & 8  & (0.4444, 0.1578) & 9  & (0.4222, 0.1367) & 10  & (0.4000, 0.0522) \\
    \midrule
    48 & 1  & (0.6000, 0.0100) & 2  & (0.5778, 0.2000) & 3  & (0.5556, 0.1156) & 4  & (0.5333, 0.0944) & 5  & (0.5111, 0.0733) \\
& 6  & (0.4889, 0.1367) & 7  & (0.4667, 0.1789) & 8  & (0.4444, 0.0522) & 9  & (0.4222, 0.0311) & 10  & (0.4000, 0.1578) \\
    \midrule
    49 & 1  & (0.6000, 0.0522) & 2  & (0.5778, 0.1367) & 3  & (0.5556, 0.1156) & 4  & (0.5333, 0.1789) & 5  & (0.5111, 0.1578) \\
& 6  & (0.4889, 0.0311) & 7  & (0.4667, 0.0733) & 8  & (0.4444, 0.2000) & 9  & (0.4222, 0.0944) & 10  & (0.4000, 0.0100) \\
    \midrule
    50 & 1  & (0.6000, 0.0100) & 2  & (0.5778, 0.0522) & 3  & (0.5556, 0.1789) & 4  & (0.5333, 0.1367) & 5  & (0.5111, 0.2000) \\
& 6  & (0.4889, 0.1156) & 7  & (0.4667, 0.1578) & 8  & (0.4444, 0.0733) & 9  & (0.4222, 0.0311) & 10  & (0.4000, 0.0944) \\
\end{longtable}
\end{landscape}

\subsection{Stopping time between RAMGapE and RA-LUCB based on $\epsilon$}\label{sec:epsilon_change}

To further investigate the comparative performance of RAMGapE against RA-LUCB in the fixed-confidence setting, we conducted additional experiments based on Experiment 1 by varying the tolerance level, $\epsilon$. 

\begin{figure}[t]
  \centering
  \subfigure[$|\mathcal{I}| = 50$, $\epsilon=0.1$]{%
    \includegraphics[width=0.48\textwidth]{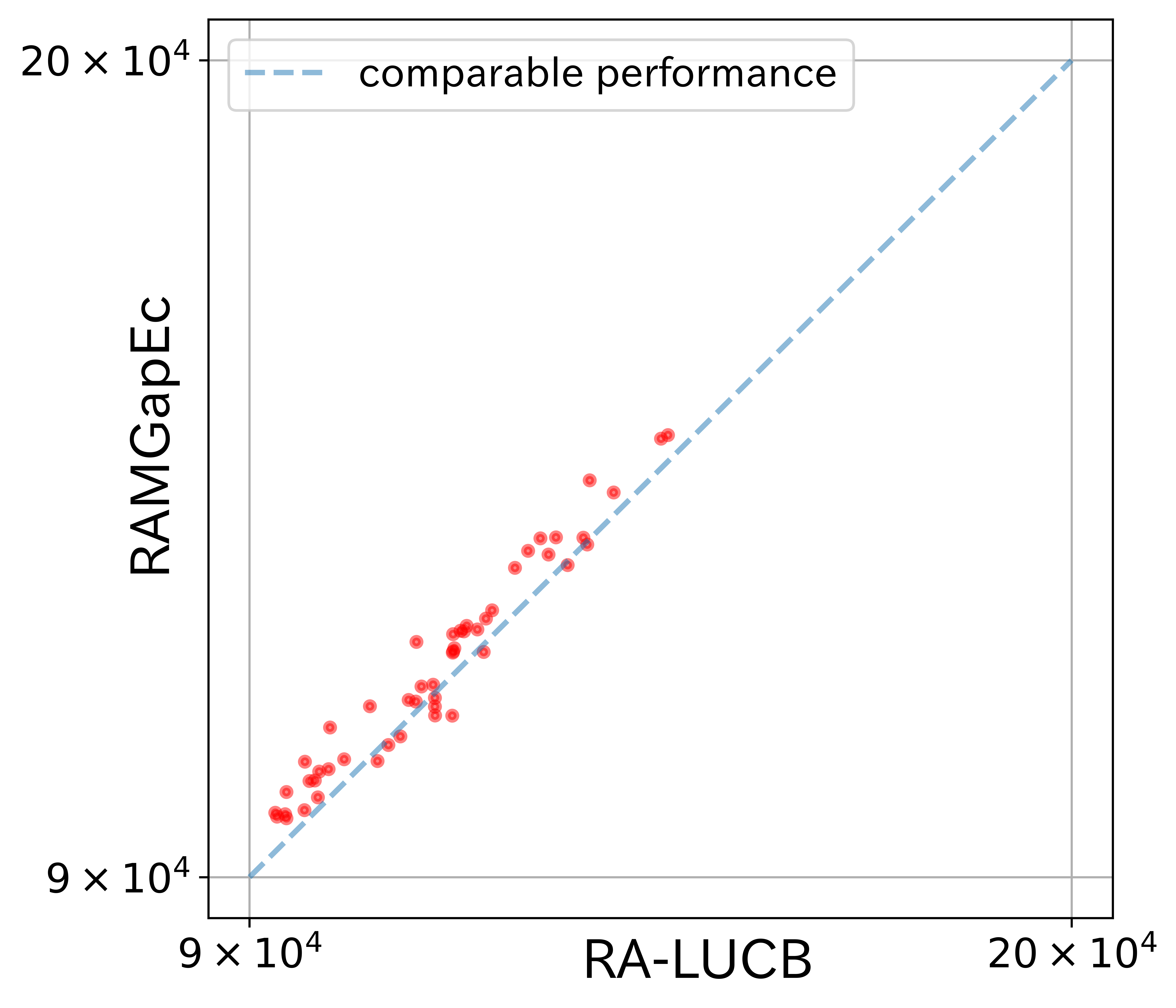}}
  \hfill
  \subfigure[$|\mathcal{I}| = 50$, $\epsilon=0.05$]{%
    \includegraphics[width=0.48\textwidth]{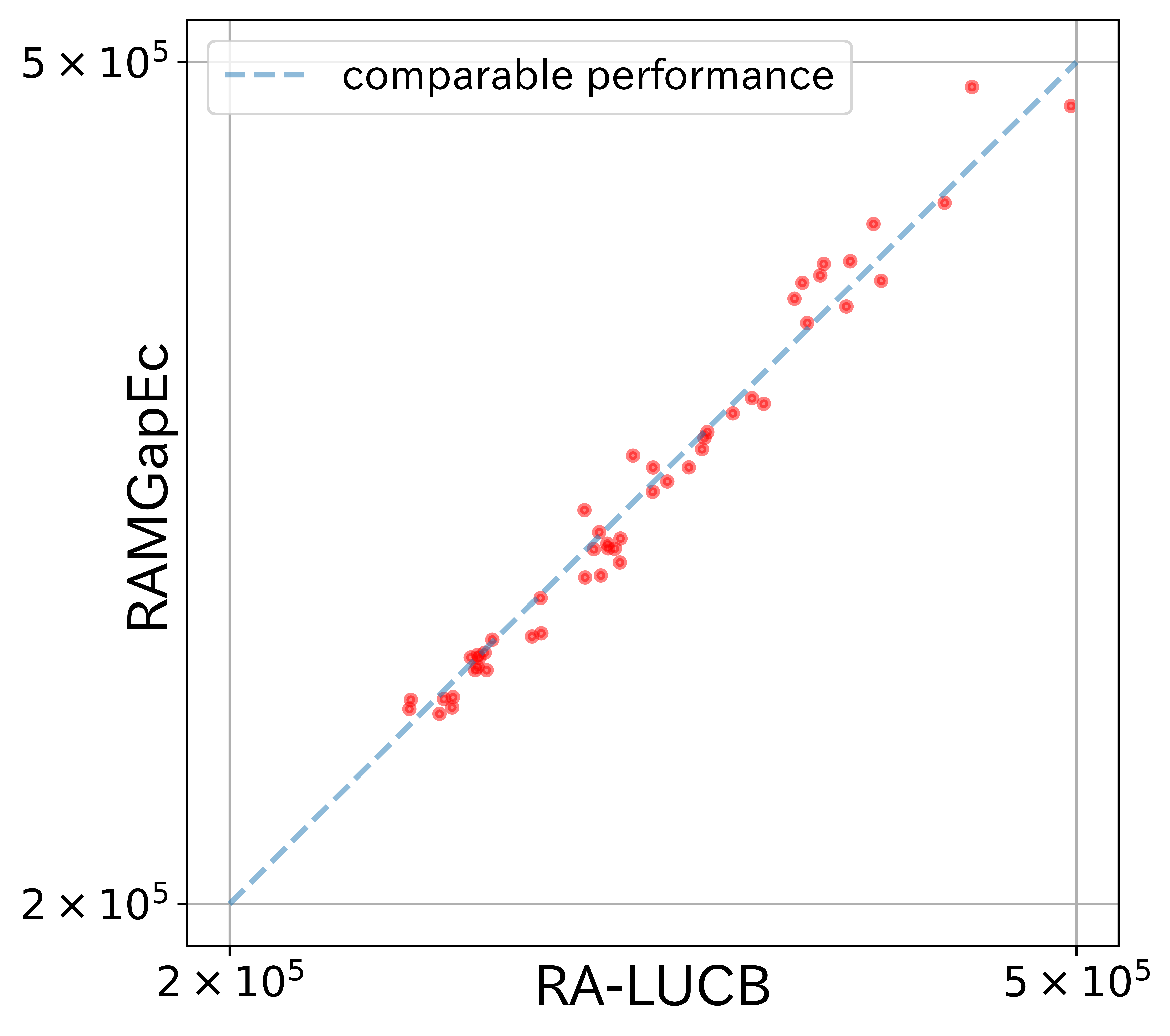}}
  \caption{\textbf{Stopping time between RAMGapE and RA-LUCB for different numbers of problem instances ($|\mathcal{I}|$) and tolerance levels ($\epsilon$).}} 

  \label{fig:complexity_analysis_fig}
\end{figure}

Fig.~\ref{fig:complexity_analysis_fig} presents scatter plots of stopping times for three settings:
\begin{itemize}
    \item[$(a)$] $|\mathcal{I}|=50, \epsilon=0.1$: This is identical to the setting in Experiment 1.
    \item[$(b)$] $|\mathcal{I}|=50, \epsilon=0.05$: We used the original 50 instances but with a smaller tolerance $\epsilon$ to create a more challenging identification task.
\end{itemize}
As shown in the plots, the points remain tightly clustered around the identity line ($y=x$) in all settings. This indicates that there are no statistically significant differences in sample efficiency between RAMGapE and RA-LUCB, even when the problem is made harder with smaller $\epsilon$. This result confirms our initial interpretation that both algorithms perform almost equally in the fixed-confidence setting we tested.

\subsection{Comparison of Pulling Ratios of Pareto and non-Pareto Arms}\label{subsec:appendix_pulling_ratios}

\begin{figure}[t]
  \centering

  \subfigure[RAMGapEb]{%
    \includegraphics[width=0.24\textwidth]{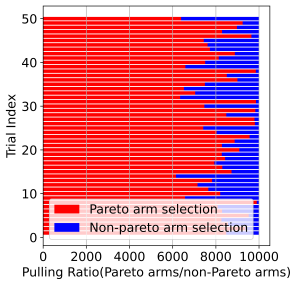}}
  \hfill
  \subfigure[RA-LUCB]{%
    \includegraphics[width=0.24\textwidth]{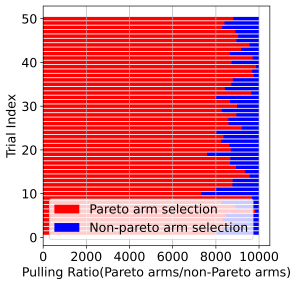}}
    \hfill
    \subfigure[LIE Round-Robin]{%
    \includegraphics[width=0.24\textwidth]{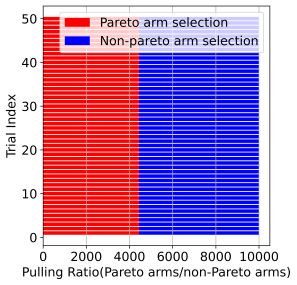}}
    \hfill
    \subfigure[Round-Robin]{%
    \includegraphics[width=0.24\textwidth]{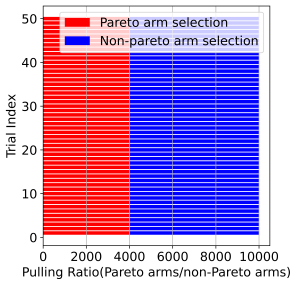}}

  \vspace{1em}

  \subfigure[EGP]{%
    \includegraphics[width=0.24\textwidth]{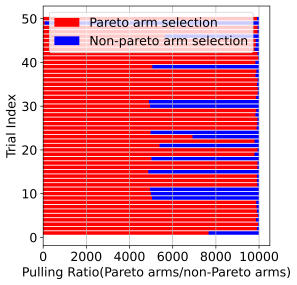}}
  \hfill
  \subfigure[$\xi$-LCB]{%
    \includegraphics[width=0.24\textwidth]{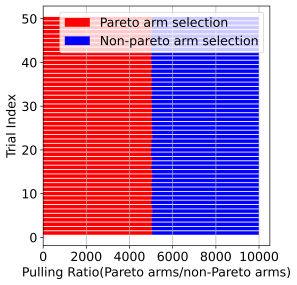}}
  \hfill
  \subfigure[HVI-Pareto]{%
    \includegraphics[width=0.24\textwidth]{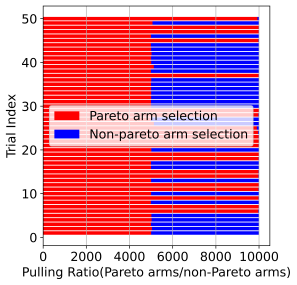}}
  \caption{\textbf{Comparison of pulling ratios of Pareto and non-Pareto arms in Experiment 3.} The vertical and horizontal axes correspond to the trial index and the pulling ratio about Pareto and non-Pareto arms per $T=10,000$, respectively. For each trial, the total number of pulling Pareto-optimal (in red), and non-Pareto arms (in blue) within $T=10,000$ is shown as bars for each method. The longer the red bars, the more frequently Pareto-optimal arms were pulled, which indicates that the algorithm focuses more effectively on the exploration and exploitation of Pareto optimal solutions.}
  \label{fig:comp_select_e3}
\end{figure}

\begin{figure}[t]
  \centering

  \subfigure[RAMGapEb]{%
    \includegraphics[width=0.24\textwidth]{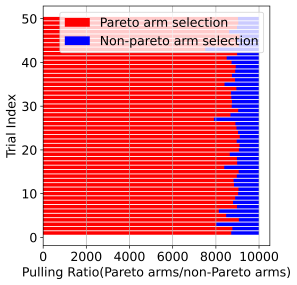}}
  \hfill
  \subfigure[RA-LUCB]{%
    \includegraphics[width=0.24\textwidth]{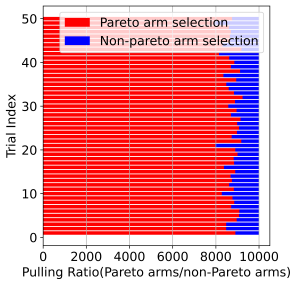}}
    \hfill
    \subfigure[LIE Round-Robin]{%
    \includegraphics[width=0.24\textwidth]{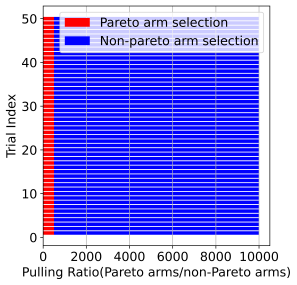}}
    \hfill
    \subfigure[Round-Robin]{%
    \includegraphics[width=0.24\textwidth]{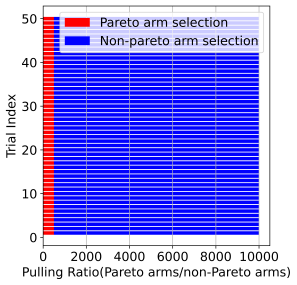}}

  \vspace{1em}

  \subfigure[EGP]{%
    \includegraphics[width=0.24\textwidth]{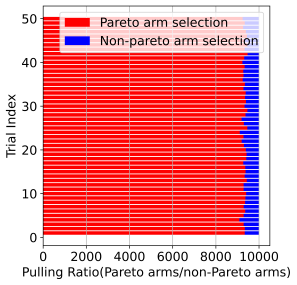}}
  \hfill
  \subfigure[$\xi$-LCB]{%
    \includegraphics[width=0.24\textwidth]{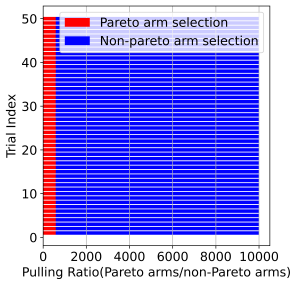}}
  \hfill
  \subfigure[HVI-Pareto]{%
    \includegraphics[width=0.24\textwidth]{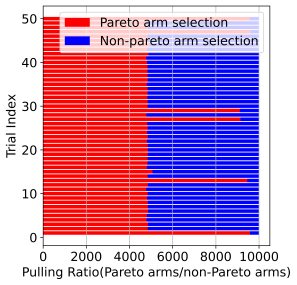}}
  \caption{\textbf{Comparison of pulling ratios for Pareto and non-Pareto arms in Experiment 4.} The meanings of the plot is the same as Fig.~\ref{fig:comp_select_e3} except $K = 100$ arms.} 
  \label{fig:comp_select_e4}
\end{figure}

To support the analysis in the main paper regarding RAMGapE's exploration strategy, this section provides a detailed visualization of the pulling ratios. Figs.~\ref{fig:comp_select_e3} and~\ref{fig:comp_select_e4} show the proportion of samples allocated to true Pareto-optimal arms (in red) versus non-Pareto arms (in blue) for each trial in Experiments 3 and 4, respectively.

These figures illustrate the outcome of RAMGapE's adaptive exploration. While other algorithms may continue to explore suboptimal arms or overly exploit a subset of arms, RAMGapE efficiently prunes non-Pareto arms. As a result, it ultimately allocates a significantly higher proportion of its budget to the Pareto set. The longer red bars for RAMGapE confirm that its strategy effectively focuses the sampling effort on the most promising regions of the decision space, which is a key factor for its strong performance.

\end{document}